\documentclass[10pt,twocolumn,letterpaper]{article}

\usepackage[pagenumbers]{cvpr} % To force page numbers, e.g. for an arXiv version

\usepackage[accsupp]{axessibility}
\usepackage{arydshln}
\usepackage{booktabs}
\usepackage{color, colortbl}
\usepackage{enumitem}
\usepackage{float}
\usepackage{hhline}
\usepackage{makecell}
\usepackage{mathrsfs}
\usepackage{multirow}
\usepackage{paralist}
\usepackage{pifont}
\usepackage{tabu}
\usepackage{threeparttable}
\usepackage{subcaption}
\usepackage{scalerel,xparse}
\usepackage{xcolor}
\usepackage{balance}
\usepackage{amsmath}
\usepackage{longtable}
\usepackage{fontawesome5}
\usepackage{xcolor}
\usepackage{graphicx}
\definecolor{gold}{rgb}{1.0, 0.84, 0.0} 
\definecolor{Gray}{gray}{0.5}
\definecolor{LGray}{gray}{0.9}
\definecolor{darkblue}{RGB}{94,110,186}
\definecolor{darkGreen}{RGB}{92, 148, 110}
\definecolor{myblue}{RGB}{14, 121, 178}
\definecolor{myred}{RGB}{192, 0, 0}
\definecolor{darkgreen}{HTML}{04bf29}
\definecolor{darkred}{HTML}{D1191F}
\definecolor{crimson}{RGB}{153, 0, 0}

\newcommand{\crimson}[1]{\textcolor{crimson}{#1}}

\newcommand{\darkblue}[1]{\textcolor{darkblue}{#1}}

\def\BenchName{SeriesBench}
\def\ApproachName{PC-DCoT}

\newcommand{\red}[1]{{\color{red}#1}}

\definecolor{cvprblue}{rgb}{0.21,0.49,0.74}
\usepackage[pagebackref,breaklinks,colorlinks,allcolors=cvprblue]{hyperref}

\def\paperID{anonymous} % *** Enter the Paper ID here
\def\confName{CVPR}
\def\confYear{2025}

\title{
SeriesBench: A Benchmark for Narrative-Driven Drama Series Understanding}

\author{Chenkai Zhang$^{1,3*}$ \quad Yiming Lei$^{1,3}$\thanks{Work done during the internship at Kuaishou Technology.} \quad Zeming Liu$^{2\dagger}$ \quad Haitao Leng$^{4\spadesuit}$  \quad Shaoguo Liu$^{4}$\\ \quad Tingting Gao$^{4}$ \quad Qingjie Liu$^{1,3\dagger}$ \quad Yunhong Wang$^{1}$\\
      $^1$State Key Laboratory of Virtual Reality Technology and Systems, Beihang University\\ 
\quad $^2$ School of Computer Science and Engineering, Beihang University, Beijing, China  \\
\quad $^3$ Hangzhou Innovation Institute, Beihang University, Hangzhou China  \\
\quad $^4$ Kuaishou Technology  \\
{\tt\small $^*$Co-first authors:~\{zhangchenkai, ymlei\}@buaa.edu.cn }\\
{\tt\small  $^\dagger$Corresponding authors $^\spadesuit$Project Leader}
}
\begin{document}
\flushbottom % 确保底部对齐
\maketitle
\begin{abstract}
% \vspace{-5pt}
With the rapid development of Multi-modal Large Language Models (MLLMs), an increasing number of benchmarks have been established to evaluate the video understanding capabilities of these models. 
However, these benchmarks focus on \textbf{standalone} videos and mainly assess ``visual elements'' like human actions and object states. In reality, contemporary videos often encompass complex and continuous narratives, typically presented as a \textbf{series}. To address this challenge, we propose \textbf{\BenchName}, a benchmark consisting of 105 carefully curated narrative-driven series, covering 28 specialized tasks that require deep narrative understanding. Specifically, we first select a diverse set of drama series spanning various genres. Then, we introduce a novel long-span narrative annotation method, combined with a full-information transformation approach to convert manual annotations into diverse task formats. To further enhance model capacity for detailed analysis of plot structures and character relationships within series, we propose a novel narrative reasoning framework, 
% \textit{i.e.}, 
\textbf{\ApproachName}. Extensive results on \textbf{\BenchName} indicate that existing MLLMs still face significant challenges in understanding narrative-driven series, while \textbf{\ApproachName} enables these MLLMs to achieve performance improvements. Overall, our \textbf{\BenchName} and \textbf{\ApproachName} highlight the critical necessity of advancing  model capabilities to understand narrative-driven series, guiding the future development of MLLMs.
SeriesBench is publicly available at \href{https://github.com/zackhxn/SeriesBench-CVPR2025}{our GitHub repository}.
% \footnote{Our dataset is publicly available at \url{https://github.com/zackhxn/SeriesBench-CVPR2025}
% }

\end{abstract}    
\vspace{-8pt}
\section{Introduction}
\label{sec:intro}

\begin{figure}[!t]
    \centering
    % \vspace{-0.3cm}
    \includegraphics[width=0.97\linewidth
    ]{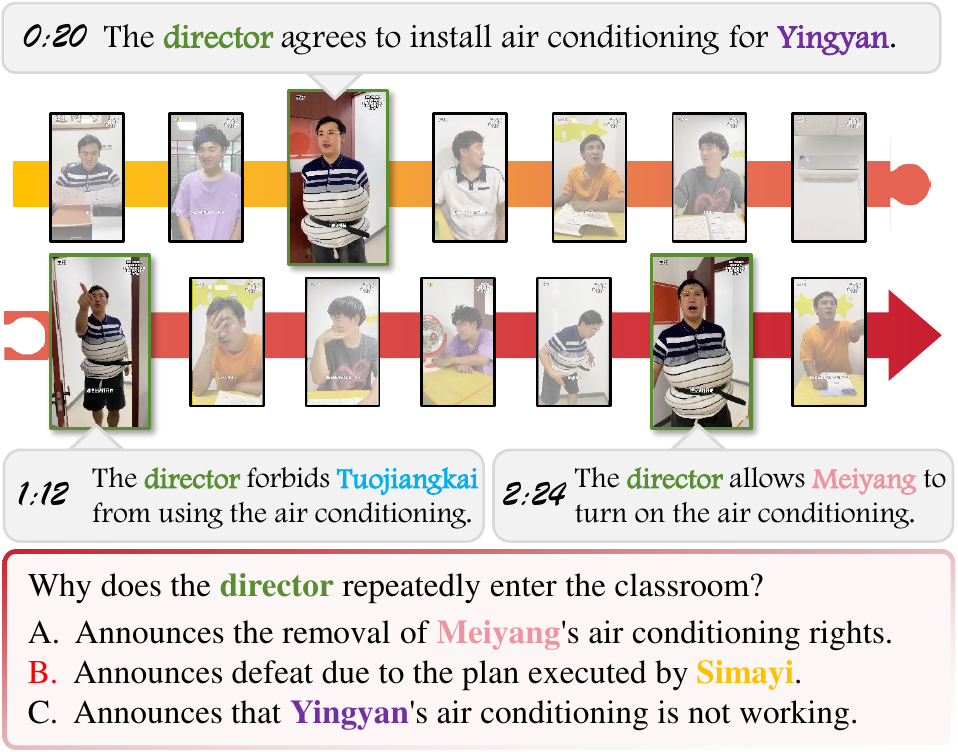}
    \vspace{-7pt}
    \caption{
        \textbf{An example from {\BenchName}.} The task involves multiple events and characters spanning a long time period in a video.
    }
    \label{fig:intro_example}
    \vspace{-0.65cm}
\end{figure}
In recent years, multimodal large language models (MLLMs) have made remarkable strides in the field of visual understanding~\cite{gpt4o_openai,gpt4o_mini_openai,vita_fu2024,videollama2_damonlpsg2024,aria_li2024,Qwen2VL,llavavideo,wang2024survey}. Building upon these advancements, a range of benchmarks~\cite{pope,xu2023lvlm,fu2023mme,mmbench,yu2023mmvetevaluatinglargemultimodal,shi2025kwaichat} have been established to evaluate the visual-language capabilities of these models, primarily through question-answering (QA) tasks based on static images. 
Although these benchmarks effectively assess static visual comprehension, real-world visual understanding must account for complex and dynamic changes over time. As a result, an increasing emphasis has emerged on evaluating MLLMs' ability to comprehend dynamic video content and perform temporal reasoning. Several studies~\cite{li2024mvbench,li2024videovista,fu2024video,liu2024etbench,liu2024tempcompass,cores2024tvbench} have introduced video-based benchmarks to assess MLLMs' temporal analysis capabilities through questions like \textit{``Is the man walking towards the right-side exit?''}.

However, despite the growing attention to temporal issues in video analysis, current evaluations \textbf{primarily focus on standalone videos}, often neglecting the narrative structure and character development across multi-video series. Furthermore, these evaluations often focus solely on visual elements, overlooking the broader multimodal nature of modern videos. In reality, most modern videos are not merely sequences of static images but rather complex multimodal compositions~\cite{FULWILER201239,mak2012visual},
integrating script, audio, and special effects through editing techniques, as seen in contemporary films and series. 
This oversight in current evaluations significantly limits model performance, as MLLMs frequently struggle with understanding intricate narrative structures and character interactions essential for comprehensive series-level video comprehension. Such limitations could impair the application of MLLMs in fields like series recommendation, interactive media, and autonomous video summarization, potentially leading to biased conclusions or skew public understanding of events, particularly in public crises or media representations.

To this end, we propose \textbf{\BenchName}, the first-ever comprehensive benchmark designed to evaluate MLLMs' understanding of narrative-driven \textbf{series}. As exemplified in Fig. \ref{fig:intro_example}, we meticulously curate \textit{105 series, comprising a total of 1,072 videos}, which span a wide variety of themes, including daily life, anime, time-travel, historical drama, fantasy, and other narrative-driven genres. Alongside the video collection, we also compiled subtitles for each video, as well as detailed information on the themes and character backgrounds specific to each series. To enable a more comprehensive evaluation of large models that aligns with the diverse modalities found in modern videos~\cite{FULWILER201239,mak2012visual}, {\BenchName} is organized into \textit{five primary task dimensions} corresponding to the core compositional elements of modern videos: \texttt{[Visuals]}, \texttt{[Script]}, \texttt{[Audio]}, \texttt{[Augmentation]}, and \texttt{[Comprehension]}, collectively covering \textit{28 fine-grained sub-tasks}. 
% 标注方法、创新的指代任务和自动QA生成
To ensure the quality of {\BenchName}, we engage \textbf{over 30 professional annotators} and propose a novel long-span narrative annotation method, which requires annotators to label narrative events and characters across extended temporal spans.
After the annotation process, we apply a full-information transformation technique to convert the annotated data into a diverse array of question types, including multiple-choice, judgment-based, and open-ended formats.

% 全面的评测和方法的创新
Inspired by human binge-watching behavior, we propose \textbf{P}lot \& \textbf{C}haracter \textbf{D}ual \textbf{C}hain of \textbf{T}hought (\textbf{\ApproachName}) to enhance MLLMs' understanding of narrative-driven series by constructing Plot Event Chains and Character Temporal Chains. 
We conduct extensive experiments and analyses on 10 leading Video-MLLMs and apply the {\ApproachName} framework to 4 top-performing models. Results show that: 1) Video-MLLMs struggle significantly with understanding narrative structure across series; 2) A notable performance gap persists between open-source and commercial MLLMs; 3) Even MLLMs with limited audio processing capabilities show partial comprehension of audio elements. 
Our evaluations demonstrate that utilizing the PC-DCoT framework can improve MLLMs' performance in narrative-driven series by over \textbf{10\%}. Despite these advancements, current MLLMs still fall short of human-level understanding of narrative-driven series, highlighting the significant potential for further research to advance MLLM series understanding.
We summarize our contributions as follows:
\\
\textbf{(1) Novel and Challenging Tasks:} We introduce a new set of challenging tasks focused on understanding video narratives within a series, involving the analysis of events and character relationships.
\\
\textbf{(2) New Video Understanding Benchmark:} We are the first to establish a video understanding benchmark specifically designed for narrative-driven series, addressing a critical gap in current evaluation benchmarks.
\\
\textbf{(3) Innovative Method and Comprehensive Evaluation:} We propose the PC-DCoT framework that improves MLLMs' capacity for understanding narrative-driven series by reconstructing inherent structure. Extensive experiments on state-of-the-art baselines and our method provide valuable insights for further research in narrative comprehension of series-based videos.

\section{Related work}
\label{sec:related_work}

\begin{table*}[!ht]
    \vspace{-5pt} % 调整间距，负值表示减少间距
    \centering
    \setlength\tabcolsep{4pt}
    \resizebox{1.0\textwidth}{!}{
        \begin{tabular}{c|c|c|l}
        \Xhline{1.0pt}
        \rowcolor{gray!20} 
        \textbf{Dimension} & \textbf{Sub-dimension} & \textbf{Fine-grained} & \textbf{Example} \\
        \Xhline{1.0pt}
        \multirow{9}{*}{\textbf{Visuals}}  
            & \multirow{3}{*}{Figures}  
            
            & \multirow{2}{*}{Actions} & \cellcolor{gray!5}{\darkblue{\textit{What was Zhao Dezhu's action when facing the room check?}}} \\ 
            & & & \cellcolor{gray!5}{\textit{(A) Hiding behind the door. \crimson{(B) Acting on the floor.} (C) Jumping out of the window.}}  \\
            
            \hhline{~|~|-|-}
            & & \multirow{1}{*}{Interactions} & \cellcolor{gray!5}{\darkblue{\textit{Was Allen asked by Xuefeng to help him take a photo?}}\ \textit{\crimson{(A) True.} (B) False.}} \\ 
            \hhline{~|-|-|-}

            & \multirow{3}{*}{Scenes}   
            
            & \multirow{1}{*}{Scene Transitions} & \cellcolor{gray!5}{\darkblue{\textit{What scenes appear in the video?}}\ \crimson{\textit{GT: Outside the office, parking lot.}}} \\ 
            
            \hhline{~|~|-|-}
            & & \multirow{2}{*}{Spatiotemporal Shifts} & \cellcolor{gray!5}{\darkblue{\textit{Where did they go if the princess's dress changed from red to white?}}} \\ 
            & & & \cellcolor{gray!5}{\textit{\crimson{(A) Demon Realm.} (B) Human Realm. (C) Magic Realm. (D) Fairy Realm.}}  \\
            \hhline{~|-|-|-}

            & \multirow{3}{*}{Objects}  
            & \multirow{1}{*}{Presence} & \cellcolor{gray!5}{\darkblue{\textit{Is the building behind the fairy's owner called Dacheng Hall?}}\ \textit{\crimson{(A) True.} (B) False.}} \\ 
            \hhline{~|~|-|-}
            
            & & \multirow{2}{*}{Interaction} & \cellcolor{gray!5}{\darkblue{\textit{What item did Manager Min hand to Xiaoqian to help take a photo?}}} \\ 
            & & & \cellcolor{gray!5}{\textit{(A) Tablet. (B) Folder. (C) Camera. \crimson{(D) Phone.}}}  \\
        \hline
        
        \multirow{17}{*}{\textbf{Script}}  
            & \multirow{3}{*}{Background} 
            
            & \multirow{2}{*}{World-Building} & \cellcolor{gray!5}{\darkblue{\textit{What type of life is described in the video?}}} \\ 
            & & & \cellcolor{gray!5}{\textit{(A) Workplace life. \crimson{(B) School life.} (C) Martial arts life. (D) Rural life.}}  \\
            \hhline{~|~|-|-}
            
            & & \multirow{1}{*}{Time and Location} & \cellcolor{gray!5}{\darkblue{\textit{According to the video, these students slept for 8 hours tonight.}}\ \textit{(A) True. \crimson{(B) False.}}} \\ 
            \hhline{~|-|-|-}

            & \multirow{10}{*}{Plot}     
            & \multirow{2}{*}{Plot Development} & \cellcolor{gray!5}{\darkblue{\textit{What was Yingyan's wish after passing the exam, and what did the director decide?}}} \\ 
            & & & \cellcolor{gray!5}{\crimson{\textit{GT: Yingyan wanted air conditioning or an eye cure and the Director agreed.}}} \\ 
            \hhline{~|~|-|-}

            & & \multirow{2}{*}{Foreshadowing and Payoff} & \cellcolor{gray!5}{\darkblue{\textit{What happened after the dorm manager praised Sima Yi's well-folded quilt?}}} \\ 
            & & & \cellcolor{gray!5}{\textit{\crimson{GT: Dorm manager was scared by spider.}}} \\ 

            \hhline{~|~|-|-}
            & & \multirow{2}{*}{Twists and Conflicts} & \cellcolor{gray!5}{\darkblue{\textit{What event marks the turning point of the plot in the story?}}} \\ 
            & & & \cellcolor{gray!5}{\textit{(A) Zhao Dezhu's appearance. (B) Director's lecture. \crimson{(C) Dorm manager's intrusion.}}} \\ 
            \hhline{~|~|-|-}

            & & \multirow{1}{*}{Climaxes and Build-ups} & \cellcolor{gray!5}{\darkblue{\textit{The chairman forgave Manager Min after expressing disappointment}}\ \textit{(A) True. \crimson{(B) False.}}} \\ 
            \hhline{~|~|-|-}

            & & \multirow{1}{*}{Suspense and Continuity} & \cellcolor{gray!5}{\darkblue{\textit{Tuojiang Kai shot at Yingyan, but Yingyan killed him instead.}} \textit{(A) True. \crimson{(B) False.}}} \\
            \hhline{~|~|-|-}

            & & \multirow{2}{*}{Emotional Dynamics} & \cellcolor{gray!5}{\darkblue{\textit{What unexpected funny moments appear in the video?}}} \\ 
            & & & \cellcolor{gray!5}{\textit{\crimson{GT: Everyone thought Yingyan was out of breath, but he could still blow music.}}}  \\
            \hhline{~|-|-|-}

            & \multirow{4}{*}{Characters} 
            & \multirow{2}{*}{Reference} & \cellcolor{gray!5}{\darkblue{\textit{Who is the person wearing a pink T-shirt and glasses?}}} \\ 
            & & & \cellcolor{gray!5}{\textit{(A) Xiaoqin. (B) Manager Huang. \crimson{(C) Brother Hong.} (D) David.}}  \\
            \hhline{~|~|-|-}
            
            & & \multirow{2}{*}{Motivations} & \cellcolor{gray!5}{\darkblue{\textit{Why did Director establish the rules for air conditioning management?}}} \\ 
            & & & \cellcolor{gray!5}{\textit{\crimson{GT: Because he did not want students to use the air conditioning easily.}}} \\ 
            \hline

        \multirow{9}{*}{\textbf{Audio}}     
            & \multirow{5}{*}{Dialogue}
            
            & \multirow{1}{*}{Dialogue Attribution} & \cellcolor{gray!5}{\darkblue{\textit{The chairman said ``My family has a daughter''.}}\ \textit{\crimson{(A) True.} (B) False.}} \\ 
           
            \hhline{~|~|-|-}
            & & \multirow{2}{*}{Pronoun References} & \cellcolor{gray!5}{\darkblue{\textit{Who does ``she'' refer to in the line ``She stole the company's confidential files''?}}} \\ 
            & & & \cellcolor{gray!5}{\textit{(A) Zhiyang. (B) Yue Xin. (C) An Xin. \crimson{(D) General Manager.}}}  \\
            \hhline{~|~|-|-}

            & & \multirow{2}{*}{Tone and Emotion} & \cellcolor{gray!5}{\darkblue{\textit{What emotions did Teacher Du display towards the top student?}}} \\ 
            & & & \cellcolor{gray!5}{\textit{(A) Relief and anger. (B) Worry and anger. (C) Anger and frustration. \crimson{(D) Relief and worry}}}  \\
            \hhline{~|-|-|-}

            & \multirow{2}{*}{Music}    
            & \multirow{2}{*}{Atmosphere} & \cellcolor{gray!5}{\darkblue{\textit{What is the main purpose of the background music in the video?}}} \\ 
            & & & \cellcolor{gray!5}{\textit{\crimson{(A) To create tension.} (B) To evoke anxiety. (C) To provide calmness. (D) To convey intensity.}}  \\
            \hhline{~|-|-|-}

            & \multirow{2}{*}{Sound Effects} 
            & \multirow{2}{*}{Impact} & \cellcolor{gray!5}{\darkblue{\textit{How do sound effects in the video express the characters' inner feelings?}}} \\ 
            & & & \cellcolor{gray!5}{\textit{\crimson{GT: They convey emotions such as anxiety and sadness through tense music and sighs.}}}  \\
        \hline

        \multirow{4}{*}{\textbf{Augmentation}} 
            & \multirow{1}{*}{Subtitles} 
           & \multirow{1}{*}{Recognition} & \cellcolor{gray!5}{\darkblue{\textit{Does the video depict the life stories of ``Shanghai boys''?}}\ \textit{\crimson{(T) True.} (F) False.}} \\
            \hhline{~|-|-|-}
            
            & \multirow{1}{*}{Labels}  
            & \multirow{1}{*}{Purpose} & \cellcolor{gray!5}{\darkblue{\textit{The prohibition label indicates no electronic devices are allowed.}} \textit{\crimson{(T) True.} (F) False.}} \\ 
            \hhline{~|-|-|-}
            
            & \multirow{2}{*}{VFX}  
            & \multirow{2}{*}{Effectiveness} & \cellcolor{gray!5}{\darkblue{\textit{Which character does Zhao Dezhu portray through the use of special effects in the play?}}} \\ 
            & & & \cellcolor{gray!5}{\textit{(A) Iron Man. \crimson{(B) Spider-Man.} (C) Hulk. (D) Thor.}}  \\
            
        \hline
        \multirow{5}{*}{\textbf{Comprehension}} 
            & \multirow{3}{*}{Engagement} 
            & \multirow{2}{*}{Future Predictions} & \cellcolor{gray!5}{\darkblue{\textit{How does the relationship between Zhiyang and Yuexin develop in the subsequent plot?}}} \\ 
            & & & \cellcolor{gray!5}{\textit{\crimson{(A) They let go of their grievances.} (B) They go their separate ways. (C) They argue again.}} 
            \\
            
            \hhline{~|~|-|-}
            & & \multirow{1}{*}{Current Interpretation} & \cellcolor{gray!5}{\darkblue{\textit{Is the little fairy struck twice, by the princess and the master?}} \cellcolor{gray!5}{\textit{\crimson{(T) True.} (F) False.}}} \\ 
            \hhline{~|-|-|-}

            & \multirow{2}{*}{Empathy} 
            & \multirow{2}{*}{Character Resonance} & \cellcolor{gray!5}{\darkblue{\textit{Which character shows the greatest passion for studying in the video?}}} \\ 
            & & & \cellcolor{gray!5}{\textit{\crimson{GT: Sima Yi demonstrates the greatest passion for learning, showing diligence and curiosity.}}}  \\
        
        \Xhline{1.0pt}
        \end{tabular}
    }
    \vspace{-6pt} % 调整间距，负值表示减少间距
    \caption{
        \textbf{Task Diversity in \BenchName.}
        The table presents different hierarchical tasks, including examples.
    }
    \vspace{-0.6cm} % 调整间距，负值表示减少间距
    \label{tab:task_dimension}
\end{table*}

\textbf{Benchmark.} 
Traditional multimodal visual benchmarks~\cite{fu2023mme,bai2023touchstone,xu2023lvlm,yu2023mmvetevaluatinglargemultimodal} primarily focused on static image understanding tasks to evaluate Image LLMs. As research shifted towards video understanding, early explorations concentrated primarily on action recognition and event detection~\cite{xu2017video,jang-IJCV-2019}, lacking the capability to comprehensively evaluate MLLMs.
Most current video benchmarks, including Video-Bench~\cite{huang2023vbench}, SEEDBench~\cite{li2024seed2plus}, and VLM-Eval~\cite{li2023vlm}, addressing more complex tasks such as video-specific understanding, knowledge-based QA, and decision-making. For long temporal sequences, benchmarks like EgoSchema~\cite{mangalam2023egoschema} and Video-MME~\cite{fu2024video} emphasize fine-grained temporal reasoning, while TempCompass~\cite{liu2024tempcompass}, MVBench~\cite{li2024mvbench}, and AutoEval-Video~\cite{chen2023autoeval} assess models with tasks demanding temporal understanding. Additionally, VideoVista~\cite{li2024videovista} and MLVU~\cite{MLVU} focus on fine-grained video understanding, with MLVU testing videos from 3 minutes to 2 hours. 
Despite their contributions to visual understanding, existing benchmarks lack attention to narrative structure \textbf{within standalone videos and across series}. 
By contrast, \textbf{\BenchName} offers a more comprehensive evaluation that emphasizes the understanding of drama series. % crafted drama series
\\
\textbf{Video-MLLM.}
The rapid development of large language models (LLMs) has accelerated multimodal research, enabling models~\cite{videochatli2024, videollavalin2024} to combine visual and textual information for a variety of tasks. With the extension of context lengths, MLLMs have been applied to video comprehension. Video-MLLMs, such as VideoChat~\cite{videochatli2024} and Video-LLaVA~\cite{videollavalin2024}, typically include a visual encoder, a projector for modal alignment, and a LLM for reasoning, significantly enhancing multimodal comprehension in complex video scenarios.
Modern Video-MLLMs, such as InternVL2~\cite{internvl2}, Qwen2-VL~\cite{Qwen2VL}, and MiniCPM-V-2.6~\cite{minicpm_yao2024}, leverage more powerful LLMs and dynamic visual encoders to improve real-time interactions and performance. VideoLLaMA2~\cite{videollama2_damonlpsg2024} uses spatial-temporal convolutions and an audio projector, expanding the integration of audio with other modalities.
However, despite these advances, most video-MLLMs still rely on sparse frame extraction within \textbf{standalone} videos, leading to challenges in narrative understanding.
Therefore, we propose a novel framework, \textbf{\ApproachName}, to improve the model's understanding of series-level narrative structures.
\section{\BenchName}
\label{sec:\BenchName}
\vspace{-5pt}
To construct {\BenchName}, we first define task dimensions based on the core elements of modern videos, as outlined in Tab. \ref{tab:task_dimension}. Then, we present our meticulously designed annotation method. Finally, we compare {\BenchName} with other benchmarks, as shown in Tab. \ref{tab:compare}.

\begin{table*}[!t]
\centering

\vspace{-5pt}
\label{T:main}
\setlength\tabcolsep{10pt}
\renewcommand{\arraystretch}{0.7} % 调小行间距（默认值为1）
\resizebox{1.0\linewidth}{!}{
\begin{tabular}{lccccccccccc}
\toprule
\multirow{3}{*}{Benchmark} & \multirow{3}{*}{Series}& \multicolumn{5}{c}{Dimension} & \multirow{3}{*}{Sub.Tokens} & \multirow{3}{*}{Anno.} & \multicolumn{2}{c}{Video}  & \multirow{3}{*}{Multi Tasks}    \\
\cmidrule(lr){3-7} \cmidrule(lr){10-11}  
 & & VS & SC & AU & AG & CO & & & \#Clips & Len.   \\
 \toprule

MSRVTT-QA~\citep{xu2017video} & \textcolor{darkred}{\ding{55}} & \textcolor{darkgreen}{\ding{51}} & \textcolor{darkred}{\ding{55}} & \textcolor{darkred}{\ding{55}} & \textcolor{darkred}{\ding{55}} & \textcolor{darkred}{\ding{55}} & \textcolor{darkred}{\ding{55}} & \includegraphics[width=0.02\textwidth]{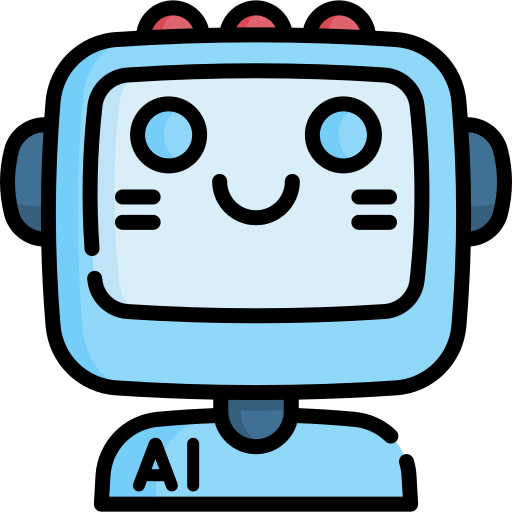}&  2,990 & 15.2 & \textcolor{darkred}{\ding{55}} \\

MSVD-QA~\citep{xu2017video} & \textcolor{darkred}{\ding{55}} & \textcolor{darkgreen}{\ding{51}} & \textcolor{darkred}{\ding{55}} & \textcolor{darkred}{\ding{55}} & \textcolor{darkred}{\ding{55}} & \textcolor{darkred}{\ding{55}} & \textcolor{darkred}{\ding{55}} & \includegraphics[width=0.02\textwidth]{figures/robot.png}&  504 & 9.8 & \textcolor{darkred}{\ding{55}} \\

TGIF-QA~\citep{jang2017tgif} & \textcolor{darkred}{\ding{55}} & \textcolor{darkgreen}{\ding{51}} & \textcolor{darkred}{\ding{55}} & \textcolor{darkred}{\ding{55}} & \textcolor{darkred}{\ding{55}} & \textcolor{darkred}{\ding{55}} & \textcolor{darkred}{\ding{55}} & \includegraphics[width=0.02\textwidth]{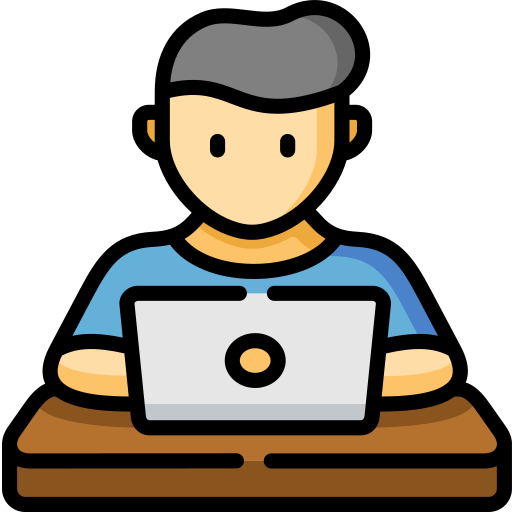} \& \includegraphics[width=0.02\textwidth]{figures/robot.png}&  9,575 & 3.0 & \textcolor{darkred}{\ding{55}} \\

ActivityNet-QA \citep{yu2019activitynet} & \textcolor{darkred}{\ding{55}} & \textcolor{darkgreen}{\ding{51}} & \textcolor{darkred}{\ding{55}} & \textcolor{darkred}{\ding{55}} & \textcolor{darkred}{\ding{55}} & \textcolor{darkred}{\ding{55}} & \textcolor{darkred}{\ding{55}} & \includegraphics[width=0.02\textwidth]{figures/human.png} &  800 & 111.4 & \textcolor{darkred}{\ding{55}} \\

% TVQA~\citep{lei2018tvqa} & \textcolor{darkred}{\ding{55}} & \textcolor{darkgreen}{\ding{51}} & \textcolor{darkred}{\ding{55}} & \textcolor{darkred}{\ding{55}} & \textcolor{darkred}{\ding{55}} & \textcolor{darkred}{\ding{55}} & 159.8 & \includegraphics[width=0.02\textwidth]{figures/human.png}&  2,179 & 11.2 & \textcolor{darkred}{\ding{55}} \\

% How2QA~\citep{li2020hero} & \textcolor{darkred}{\ding{55}} & \textcolor{darkgreen}{\ding{51}} & \textcolor{darkred}{\ding{55}} & \textcolor{darkred}{\ding{55}} & \textcolor{darkred}{\ding{55}} & \textcolor{darkred}{\ding{55}} & 31.1 & \includegraphics[width=0.02\textwidth]{figures/human.png} & 1,166 & 15.3 & \textcolor{darkred}{\ding{55}} \\

STAR~\citep{wu2021star_situated_reasoning} & \textcolor{darkred}{\ding{55}} & \textcolor{darkgreen}{\ding{51}} & \textcolor{darkred}{\ding{55}} & \textcolor{darkred}{\ding{55}} & \textcolor{darkred}{\ding{55}} & \textcolor{darkred}{\ding{55}} & \textcolor{darkred}{\ding{55}} & \includegraphics[width=0.02\textwidth]{figures/robot.png}&  914 & 11.9 & \textcolor{darkred}{\ding{55}} \\

NExT-QA~\citep{xiao2021next} & \textcolor{darkred}{\ding{55}} & \textcolor{darkgreen}{\ding{51}} & \textcolor{darkred}{\ding{55}} & \textcolor{darkred}{\ding{55}} & \textcolor{darkred}{\ding{55}} & \textcolor{darkred}{\ding{55}} & \textcolor{darkred}{\ding{55}} & \includegraphics[width=0.02\textwidth]{figures/robot.png}& 1,000 & 39.5 & \textcolor{darkred}{\ding{55}} \\

\midrule
MVBench~\citep{li2024mvbench} & \textcolor{darkred}{\ding{55}} & \textcolor{darkgreen}{\ding{51}} & \textcolor{darkred}{\ding{55}} & \textcolor{darkred}{\ding{55}} & \textcolor{darkred}{\ding{55}} & \textcolor{darkred}{\ding{55}} & \textcolor{darkred}{\ding{55}} & \includegraphics[width=0.02\textwidth]{figures/robot.png}& 3,641 & 16.0 & \textcolor{darkred}{\ding{55}} \\

Video-Bench~\citep{ning2023video} & \textcolor{darkred}{\ding{55}} & \textcolor{darkgreen}{\ding{51}} & \textcolor{darkred}{\ding{55}} & \textcolor{darkred}{\ding{55}} & \textcolor{darkred}{\ding{55}} & \textcolor{darkred}{\ding{55}} & \textcolor{darkred}{\ding{55}} & \includegraphics[width=0.02\textwidth]{figures/human.png} \& \includegraphics[width=0.02\textwidth]{figures/robot.png}&  5,917 & 56.0 & \textcolor{darkred}{\ding{55}} \\

EgoSchema~\citep{mangalam2024egoschema} & \textcolor{darkred}{\ding{55}} & \textcolor{darkgreen}{\ding{51}} & \textcolor{darkred}{\ding{55}} & \textcolor{darkred}{\ding{55}} & \textcolor{darkred}{\ding{55}} & \textcolor{darkred}{\ding{55}} & \textcolor{darkred}{\ding{55}} & \includegraphics[width=0.02\textwidth]{figures/human.png} \& \includegraphics[width=0.02\textwidth]{figures/robot.png}&  5,063 & 180.0 & \textcolor{darkred}{\ding{55}} \\

AutoEval-Video~\citep{chen2023autoeval} & \textcolor{darkred}{\ding{55}} & \textcolor{darkgreen}{\ding{51}} & \textcolor{darkred}{\ding{55}} & \textcolor{darkred}{\ding{55}} & \textcolor{darkred}{\ding{55}} & \textcolor{darkred}{\ding{55}} & \textcolor{darkred}{\ding{55}} & \includegraphics[width=0.02\textwidth]{figures/human.png} &  327 & 14.6 & \textcolor{darkred}{\ding{55}} \\

TempCompass~\citep{Liu2024TempCompassDV} & \textcolor{darkred}{\ding{55}} & \textcolor{darkgreen}{\ding{51}} & \textcolor{darkred}{\ding{55}} & \textcolor{darkred}{\ding{55}} & \textcolor{darkred}{\ding{55}} & \textcolor{darkred}{\ding{55}} & \textcolor{darkred}{\ding{55}} & \includegraphics[width=0.02\textwidth]{figures/human.png} \& \includegraphics[width=0.02\textwidth]{figures/robot.png}& 410 & 11.4 & \textcolor{darkred}{\ding{55}} \\

Video-MME-S~\cite{fu2024video} & \textcolor{darkred}{\ding{55}} & \textcolor{darkgreen}{\ding{51}} & \textcolor{darkred}{\ding{55}} & \textcolor{darkred}{\ding{55}} & \textcolor{darkred}{\ding{55}} & \textcolor{darkred}{\ding{55}} & 198.6 & \includegraphics[width=0.02\textwidth]{figures/human.png}&  300 & 80.7 & \textcolor{darkred}{\ding{55}} \\

VideoVista~\cite{li2024videovista} & \textcolor{darkred}{\ding{55}} & \textcolor{darkgreen}{\ding{51}} & \textcolor{darkred}{\ding{55}} & \textcolor{darkred}{\ding{55}} & \textcolor{darkred}{\ding{55}} & \textcolor{darkred}{\ding{55}} & \textcolor{darkred}{\ding{55}} & \includegraphics[width=0.02\textwidth]{figures/robot.png}&  3402 & 131.0 & \textcolor{darkred}{\ding{55}} \\

TVBench~\cite{cores2024tvbench} & \textcolor{darkred}{\ding{55}} & \textcolor{darkgreen}{\ding{51}} & \textcolor{darkred}{\ding{55}} & \textcolor{darkred}{\ding{55}} & \textcolor{darkred}{\ding{55}} & \textcolor{darkred}{\ding{55}} & \textcolor{darkred}{\ding{55}} & \includegraphics[width=0.02\textwidth]{figures/robot.png} &  2525 & 446.1 & \textcolor{darkred}{\ding{55}} \\

MMBench-Video~\cite{fang2024mmbenchvideo} & \textcolor{darkred}{\ding{55}} & \textcolor{darkgreen}{\ding{51}} & \textcolor{darkred}{\ding{55}} & \textcolor{darkred}{\ding{55}} & \textcolor{darkred}{\ding{55}} & \textcolor{darkred}{\ding{55}} & \textcolor{darkred}{\ding{55}} & \includegraphics[width=0.02\textwidth]{figures/human.png}&  609 & 165.0 & \textcolor{darkred}{\ding{55}} \\

\midrule
\textbf{\BenchName\ (Ours)} & \textcolor{darkgreen}{\ding{51}} & \textcolor{darkgreen}{\ding{51}} & \textcolor{darkgreen}{\ding{51}} & \textcolor{darkgreen}{\ding{51}} & \textcolor{darkgreen}{\ding{51}} & \textcolor{darkgreen}{\ding{51}} & \textbf{250.2} &  \includegraphics[width=0.02\textwidth]{figures/human.png} & 1072 & 79.2 &\textcolor{darkgreen}{\ding{51}} \\
\bottomrule 

\end{tabular}}
\caption{
    \textbf{Comparison between  SeriesBench and other existing benchmarks.}
    Compared to other benchmarks, \BenchName\ extends video understanding to narrative-driven series, while other benchmarks are limited to the visual comprehension of standalone videos. In the table, 
    \textbf{Series} indicates whether the dataset contains episodic or series-based videos.
    \textbf{VS} represents visuals, \textbf{SC} represents script, \textbf{AU} represents audio, \textbf{AG} represents augmentation, and \textbf{CO} represents comprehension.
    \textbf{Sub.Tokens} refers to the average number of subtitle tokens per video.
    \includegraphics[width=0.02\textwidth]{figures/human.png} denotes manual annotation, and \includegraphics[width=0.02\textwidth]{figures/robot.png} represents automatic annotation.
    % \textbf{Chr.Anno} indicates whether characters within the video are annotated with names and portraits.
    \textbf{\#Clips} indicates the number of video clips, \textbf{Len.} is the average video length. 
    \textbf{Multi Tasks} reflects whether the benchmark includes a variety of task types.
}
\vspace{-0.5cm}
\label{tab:compare}
\end{table*}

% 任务维度讲解
\subsection{Task Dimension Definitions}
% \balance % 在这一段前添加 \balance
Existing benchmarks for evaluating MLLMs are largely confined to purely visual elements and often rely on standalone video clips, limiting their capacity to capture complex and cohesive narratives. For instance, these benchmarks typically assess continuous image frames that depict human actions or object states, such as asking, \textit{``What object is this person passing?''} However, they rarely delve into deeper narrative questions like \textit{``Why is this person passing this object?''} and fail to explore broader narrative reasoning across connected videos. 
To address this challenge, {\BenchName} offers a more comprehensive evaluation, with a strong emphasis on narrative understanding by utilizing videos from drama series. To align with the diverse elements inherent in modern videos, which include frames, scripts, audio, and post-production components, we have meticulously designed five major task categories for \BenchName: \texttt{[Visuals]}, analyzing frames and figure appearances critical to the narrative; \texttt{[Script]}, focusing on the design and development of plot and character dynamics; 
\texttt{[Audio]}, interpreting dialogue, sound effects, and music in context; \texttt{[Augmentation]}, examining special effects and labels affecting narrative; and \texttt{[Comprehension]}, integrating all elements to assess overall narrative comprehension.
All tasks are aimed at enhancing the understanding of narrative-driven series, with examples provided in Tab. \ref{tab:task_dimension}.

% 人工标注，指代标注，QA转换
\subsection{Task Annotation Design}
The construction of {\BenchName} is a meticulous process designed to capture the complexity of a narrative-driven series. We introduce a novel long-span narrative annotation method, requiring annotators to label events and characters across extended temporal spans. 
% Additionally, considering the poor performance of MLLMs in character reasoning, we propose a two-tier character coreference task to further evaluate the model's capacity to identify and link character references.
To further support task generation, we apply a full-information transformation method to convert annotations into diverse questions, enhancing the benchmark's quality and challenge.
% The complete pipeline is shown in Fig. ???\ref{fig:pipeline}.

\vspace{-7pt}
\subsubsection{Long-span Narrative Annotation}
After collecting the raw series, we established detailed annotation guidelines and trained professional annotators to carry out the video annotations. We also conducted a trial annotation phase and selected the 32 annotators with the highest annotation quality.
First, annotators were required to fully understand the video narrative before identifying key segments, including major events and the actions of key characters.
Next, they summarized the annotations into a declarative statement that integrated all relevant content. These statements typically described long-term events or character developments and often involved narrative reasoning. Annotators then selected the task dimension most relevant to each statement. 
Finally, all annotated content underwent a rigorous quality control process conducted by dedicated reviewers. Each annotation was meticulously examined for accuracy, coherence, and consistency. If any annotation failed to meet the established quality standards, it was sent back for correction and refinement. Ultimately, a random sample of 500 annotations showed that 96\% met the required standards, demonstrating a high level of quality.
% This annotation workflow ensures that each statement covers information spanning a long duration in the video and incorporates multi-hop narrative reasoning content, thereby guaranteeing both the quality of the annotations and the complexity required for subsequent question generation.
This workflow ensures that each statement captures extended narrative content and supports complex, multi-hop reasoning—laying a solid foundation for subsequent question generation.

\vspace{-7pt}
\subsubsection{Full-Information Transformation}
\vspace{-3pt}
\begin{figure*}[!ht]
    \centering
    % \vspace{-0.3cm}
    \includegraphics[width=1.0\linewidth
    ]{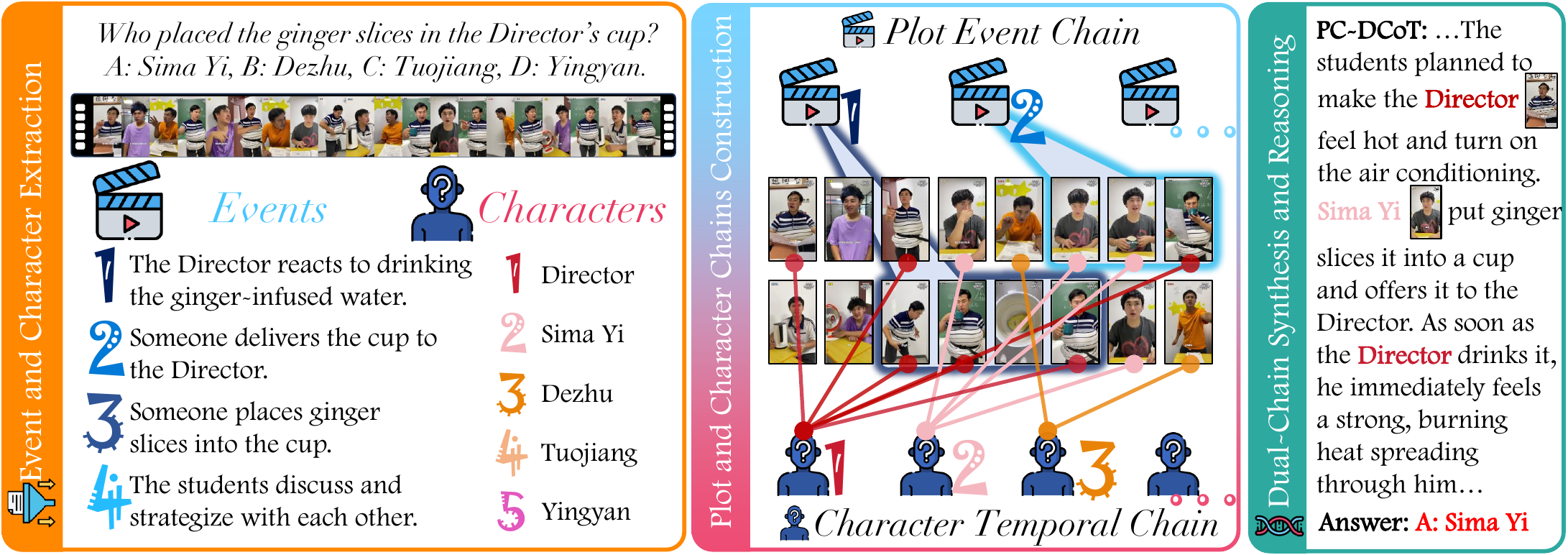}
    \vspace{-15pt}
    \caption{
        \textbf{Illustration of PC-DCoT.}
        The process begins by extracting relevant events and characters from the input question and the original video. Subsequently, a Plot Event Chain and a Character Temporal Chain are constructed independently. Finally, these chains are merged, enabling a reasoning process that generates an answer to the posed question based on the integrated dual-chain framework.
    }
    \label{fig:PC-DCoT}
    \vspace{-0.5cm}
\end{figure*}

Based on the annotated data, we proposed a full-information transformation method to generate tasks. Due to each video annotation containing manual declarative statements,  subtitles, themes, and character introductions gathered during the video collection phase, we can utilize the video's full information to transform the manual annotations effectively. Leveraging this data, we use GPT-4o to generate questions of various task types, including true-or-false questions, multiple-choice questions, and open-ended questions. This method ensures that annotated declarative sentences are seamlessly transformed into question stems and correct answers, with related video information crafted as distractors. Finally, A total of 29,196 tasks were generated. Specific examples can be found in Tab. \ref{tab:task_dimension}. 

\subsection{Comparison with Previous Benchmarks}
Previous benchmarks primarily focus on standalone videos and evaluate only the visual elements within them. In comparison, as shown in Tab. \ref{tab:compare}, videos in {\BenchName} are organized as interconnected series collections, allowing for a more comprehensive evaluation of narrative-driven content.
To achieve this, we crawled the top 11 trending series genres from \textbf{\textit{Kuaishou}}\footnote{https://www.kuaishou.cn}, 
and through data cleaning, we finalized a selection of 105 series with the highest engagement ratios.
Furthermore, {\BenchName} expands task dimensions beyond visual elements to include narrative, audio, augmentation, and comprehension, promoting a more comprehensive understanding of the videos. As these videos encompass complete narratives, they contain a higher volume of subtitles. 
To further enhance the quality, we enlisted over 30 professional annotators to meticulously label the videos, supported by a rigorous quality review mechanism.
In addition to various new features, {\BenchName} comprises 1,072 narrative-driven videos, and diverse task types, enabling comprehensive evaluation of video understanding.

\section{Plot \& Character Dual Chain of Thought}
\label{sec:PC-DCoT}

Current MLLMs~\cite{gpt4o_openai,videollama2_damonlpsg2024,Qwen2VL,internvl2} are capable of describing character actions and object states in a video but struggle to comprehend narrative-related events, character identities, and inter-character relationships. However, humans naturally focus on the distinct identities and roles of characters and the complex events unfolding between them when watching narrative-driven videos, particularly in drama series. Drawing inspiration from this human approach, we propose the \textbf{PC-DCoT} framework. 
This framework begins by extracting key events and characters that need to be tracked based on the given video and questions, then constructs separate analytical chains for events and characters, which are subsequently integrated into a dual-chain structure to facilitate comprehensive answer reasoning.
Fig. \ref{fig:PC-DCoT} illustrates the overall framework of PC-DCoT.  

% 事件与人物拆解
\subsection{Event and Character Extraction.}
For narrative-driven videos, the plot and characters are critical components that drive the storyline. To effectively analyze this complex content, the method begins by leveraging the MLLM to receive raw video frame input, along with relevant questions. Then, the MLLM extracts and identifies significant events pertinent to the questions and key characters that need to be tracked. This comprehensive process establishes primary targets for tracking, thereby laying the groundwork for a deeper and more comprehensive understanding of the narrative.
% 分别构建剧情事件链与人物时间链

\subsection{Plot and Character Chains Construction.}
In narrative-driven videos, character appearances are often discontinuous, while events typically progress through interconnected and cohesive segments.
To align input videos with task-relevant events and characters, we trained a video clip model~\cite{radford2021learningtransferablevisualmodels} to identify and isolate key scenes. By taking the raw video input along with extracted events relevant to the questions, the model searches for video frames that correspond to each identified event. It then aggregates these frames with adjacent frames to form cohesive sets of independent event sequences. Subsequently, the MLLM is employed to generate detailed descriptions of each continuous event, constructing the Plot Event Chain. Simultaneously, the model utilizes key character portraits to identify and retrieve all corresponding character frames from the video. By recognizing and tracking the appearances of these characters, even when their presence is disjointed and spread across different segments, the MLLM generates detailed behavioral descriptions for each individual. This approach ensures consistent tracking of specific characters throughout the entire video, resulting in the construction of multiple Character Temporal Chains that capture their behaviors, interactions, and roles over the video timeline.
% 合成双链推理回答
\subsection{Dual-Chain Synthesis and Reasoning.}
Since the Plot Event Chain and Character Temporal Chain both contain specific video frames, each event or character is associated with a precise time interval. This precise temporal annotation allows for direct and accurate alignment of events and character appearances based on their temporal axes, making it possible to determine exactly which characters are involved within the timeframe of each event. Subsequently, the MLLM synthesizes the interactions between events and characters, merging their occurrences and behaviors within the aligned timeframes. This integration yields a more cohesive and detailed narrative representation, capturing the nuances of character dynamics, interactions, and event progression with enhanced clarity.

\section{Experiments}
\label{sec:experiments}

\subsection{Settings}
We partitioned {\BenchName} into training, validation, and test sets using an 8:1:1 ratio through stratified random sampling from the original dataset, ensuring no overlap between different sets. For multiple-choice and judgment tasks, we evaluated the correctness of discrete predictions using accuracy. For open-ended tasks, we measured lexical overlap and semantic relevance using BLEU-2~\cite{papineni2002bleu} and METEOR~\cite{banerjee2005meteor}, and BERTScore F1~\cite{bert-score} for semantic similarity.

We evaluate a comprehensive set of recent Video-MLLMs following their official inference and frame extraction configurations similar to Video-MME-S~\cite{fu2024video} and VideoVista~\cite{li2024videovista}: (a) Open sourced \textit{Video}-MLLMs including InternVL2~\cite{internvl2}, LLaVA-OneVision~\cite{llavaov}, Qwen2-VL~\cite{Qwen2VL}, LLaVA-Video~\cite{llavavideo}, MiniCPM-V 2.6~\cite{minicpm_yao2024}, Aria~\cite{aria_li2024}, and VITA~\cite{vita_fu2024}. (b)Open sourced \textit{Video-Audio}-MLLMs, \textit{i.e.}, VideoLLaMA2.1-AV~\cite{videollama2_damonlpsg2024}. (c) Proprietary commercial models such as GPT-4o-mini~\cite{gpt4o_mini_openai} and GPT-4o~\cite{gpt4o_openai}. 

\subsection{Main Results on \BenchName}
\begin{table*}[!ht]
    \centering
    \vspace{-5pt}
    \small
    \renewcommand\tabcolsep{8pt} % column space
    \renewcommand\arraystretch{0.85} % row space
    \resizebox{1.0\linewidth}{!}{
        \begin{tabular}{l|c|c|c|c|c|c|c|c|c|c|c}
            \toprule
            \rowcolor{gray!40} 
            \multicolumn{1}{c|}{\textbf{Model}}
            & \textbf{Size}
            % & \makecell{\textbf{LLM} \\ \textbf{Params}}
            & \textbf{Frames}
            & \textbf{VS}
            & \textbf{SC}
            & \textbf{AU}
            & \textbf{AG}
            & \textbf{CO}
            & \textbf{Overall}
            & \textbf{BL-2}
            & \textbf{MET}
            & \textbf{F1$_{\text{BERT}}$} \\
            \midrule
            \rowcolor{gray!20} 
            \multicolumn{12}{c}{\textit{Heuristic baselines}} \\
            \midrule
            Random Choice & - & - & 39.3 & 38.2 & 35.5 & 36.5 & 38.8 & \textbf{37.7} & - & - & - \\
            Frequent Guess & - & - & 43.6 & 46.0 & 43.1 & 41.1 & 50.6 & \textbf{44.4} & - & - & - \\
            Human & - & - & 98.2 & 94.4 & 94.6 & 97.2 & 92.6 & \textbf{95.8} & - & - & - \\

            \midrule
            \rowcolor{gray!20} 
            \multicolumn{12}{c}{\textit{Open-source Video MLLMs}} \\
            \midrule
            
            InternVL2\cite{internvl2} & 7B & 32 
            & 52.0 & 56.8 & \cellcolor{gray!30}{57.7} & \cellcolor{orange!25}{78.6} & 57.7 & \cellcolor{gray!30}{\textbf{59.2}} & 8.43 & 21.22 & 68.40 \\

            LLaVA-OneVision\cite{llavaov} & 7B & 32 
            & 51.5 & 54.0 & 56.2 & 70.6 & \cellcolor{purple!30}{59.6} & \textbf{56.9} & 7.30 & 20.20 & 67.82 \\

            LLaVA-Video\cite{llavavideo} & 7B & 64 
            & \cellcolor{gray!30}{54.9} & 56.8 & 57.4 & 71.2 & 51.9 & \textbf{58.3} & 7.27 & 19.06 & 67.95 \\

            Qwen2-VL\cite{Qwen2VL} & 7B & 64 
            & \cellcolor{orange!25}{55.7} & \cellcolor{orange!25}{57.5} & \cellcolor{orange!25}{58.6} & 75.3 & \cellcolor{purple!30}{59.6} & \cellcolor{orange!25}{\textbf{60.3}} & \cellcolor{purple!30}{11.41} & \cellcolor{gray!30}{27.97} & \cellcolor{purple!25}{70.71} \\

            MiniCPM-V 2.6\cite{minicpm_yao2024} & 8B & 64 
            & 53.3 & \cellcolor{gray!30}{57.3} & \cellcolor{gray!30}{57.7} & \cellcolor{gray!30}{76.6} & 55.1 & \textbf{59.1} & 8.57 & \cellcolor{purple!30}{30.06} & 68.66 \\

            Aria\cite{aria_li2024} & 8×3.5B & 128 
            & 52.4 & 56.6 & 51.8 & 75.6 & 50.0 & \textbf{56.8} & \cellcolor{orange!25}{10.00} & 26.08 & \cellcolor{orange!25}{70.35} \\

            VITA\cite{vita_fu2024} & 8×7B & 32 
            & 46.9 & 46.0 & 53.3 & 68.9 & 48.1 & \textbf{51.8} & 9.32 & 27.08 & 68.93 \\

            \midrule
            \rowcolor{gray!20} 
            \multicolumn{12}{c}{\textit{Open-source Video-Audio MLLMs}} \\
            \midrule

            VideoLLaMA2.1-AV\cite{videollama2_damonlpsg2024} & \multirow{2}{*}{7B} & \multirow{2}{*}{32} 
            & 47.4 & 52.2 & 53.3 & 66.6 & 49.4 & \textbf{53.1} & 7.61 & 23.34 & 67.53 \\
            \multicolumn{1}{r|}{+audio} & & 
            & 45.3 & 52.2 & 51.3 & 66.2 & 46.2 & \textbf{51.7} & 7.24 & 22.71 & 67.09 \\
            
            \midrule
            \rowcolor{gray!20} 
            \multicolumn{12}{c}{\textit{Commercial MLLMs}} \\
            \midrule

            GPT-4o-mini\cite{gpt4o_mini_openai} & N/A & 50 
            & 46.7 & 42.7 & 47.9 & 70.9 & 44.9 & \textbf{49.8} & \cellcolor{gray!30}{9.99} & \cellcolor{orange!25}{29.30} & 68.76 \\
            
            GPT-4o\cite{gpt4o_openai} & N/A & 50 
            & \cellcolor{purple!30}{55.8} & \cellcolor{purple!30}{62.8} & \cellcolor{purple!30}{60.6} & \cellcolor{purple!30}{79.9} & \cellcolor{purple!30}{59.6} & \cellcolor{purple!30}{\textbf{62.8}} & 9.61 & 25.10 & \cellcolor{gray!30}{68.94}\\
            
            \midrule
            \rowcolor{gray!20} 
            \multicolumn{12}{c}{\textit{MLLMs with \ApproachName}} \\
            \midrule
            
            InternVL2\red{{$^\dag$}} & 7B & -
            % & 76.5 {\footnotesize (\textit{+24.5})} & 71.4 {\footnotesize (\textit{+14.6})} & 67.3 {\footnotesize (\textit{+9.6})} & 81.1 {\footnotesize (\textit{+2.5})} & 66.7 {\footnotesize (\textit{+9.0})} & \textbf{73.3} {\footnotesize (\textit{+14.1})} & 8.26 & 20.71 & 26.24 \\
            & 76.5 & 71.4 & 67.3 & 81.1 & 66.7 & \textbf{73.3} {\footnotesize (\textit{{$\uparrow$}14.1})} & 8.26 & 20.71 & 67.85 \\
            
            Qwen2-VL\red{{$^\dag$}} & 7B & -
            % & 75.2 {\footnotesize (\textit{+19.5})} & 73.5 {\footnotesize (\textit{+16.0})} & 69.7 {\footnotesize (\textit{+11.1})} & 80.9 {\footnotesize (\textit{+5.6})} & 67.4 {\footnotesize (\textit{+7.8})} & \textbf{73.9 }{\footnotesize (\textit{+13.6})} & 8.93 & 23.41 & 27.93 \\
            & 75.2 & 73.5 & 69.7 & 80.9 & 67.4 & \textbf{73.9 }{\footnotesize (\textit{{$\uparrow$}13.6})} & 8.93 & 23.41 & 68.08 \\

            MiniCPM-V 2.6\red{{$^\dag$}} & 8B & - 
            % & 76.3 {\footnotesize (\textit{+23.0})} & 68.2 {\footnotesize (\textit{+10.9})} & 66.5 {\footnotesize (\textit{+8.8})} & 82.5 {\footnotesize (\textit{+5.9})} & 60.0 {\footnotesize (\textit{+4.9})} & \textbf{72.0} {\footnotesize (\textit{+12.9})} & 14.16 & 33.64 & 36.00 \\
            & 76.3 & 68.2 & 66.5 & 82.5 & 60.0 & \textbf{72.0} {\footnotesize (\textit{{$\uparrow$}12.9})} & 14.16 & 33.64 & 72.98 \\

            GPT-4o\red{{$^\dag$}} & N/A & -
            % & 78.6 {\footnotesize (\textit{+22.8})} & 76.1 {\footnotesize (\textit{+13.3})} & 73.8 {\footnotesize (\textit{+13.2})} & 82.1 {\footnotesize (\textit{+2.2})} & 61.7 {\footnotesize (\textit{+2.1})} & \textbf{76.2} {\footnotesize (\textit{+13.4})} & 12.61 & 28.29 & 33.86 \\
            & 78.6 & 76.1 & 73.8 & 82.1 & 61.7 & \textbf{76.2} {\footnotesize (\textit{{$\uparrow$}13.4})} & 12.61 & 28.29 & 72.75 \\

            \bottomrule
        \end{tabular}
    }
    
    \vspace{-5pt}
    \caption{
        \textbf{Performance of MLLMs on {\BenchName}.}
        Size means the LLM size.
        Judgement and multichoice metrics Accuracy and open-ended metrics BLEU-2(\textbf{BL-2}), METEOR (\textbf{MET}), and BERTScore F1 (\textbf{F1$_{\text{BERT}}$}) are reported in percentage (\%).
        % Overall evaluation scores(\textbf{Overall}) are calculated by averaging the scores of judgment and multichoice metrics.
        ``-'' indicates that results are not feasible with open-ended metrics in heuristic baselines.
        \red{{$\dag$}}: MLLMs using the {\ApproachName} framework during inference.
        The best, second-best, and third-best results are marked \colorbox{purple!30}{purple}, \colorbox{orange!25}{orange}, and \colorbox{gray!30}{gray}, respectively.
    }
    \vspace{-0.5cm} % 调整间距，负值表示减少间距
    \label{tab:main_results}
\end{table*}

% \cellcolor{purple!30}    % top 1 - 浅紫色
% \cellcolor{orange!25}    % top 2 - 浅橙色
% \cellcolor{gray!30}      % top 3 - 浅灰色

We include random choice and frequent guessing as naive baselines, alongside human performance as the upper bound. The overall evaluation results averaged across sub-tasks, are presented in Tab.~\ref{tab:main_results}. We further apply {\ApproachName} to the models that performed best, specifically InternVL2, Qwen2-VL, MiniCPM-V 2.6, and GPT-4.

\textbf{Strong Video-LLMs on existing benchmarks struggle on {\BenchName}.}
While state-of-the-art(SOTA) MLLMs (e.g., GPT-4o and Qwen2-VL) have achieved around 80\% accuracy on benchmarks(e.g., Video-MME-S and VideoVista), their performance on {\BenchName} remains limited, particularly for tasks that require fine-grained visual analysis and deep comprehension, which demands enhanced content understanding and reasoning. We attribute the models' inefficiencies to disrupted visual continuity caused by discrete frame sampling and an overemphasis on image-level captioning rather than on narrative storyline in episodes.

\textbf{Performance gap between open-source and commercial MLLMs remains evident.} 
We observe that GPT-4o achieves the highest performance on {\BenchName}, particularly in \texttt{[Script]} tasks involving background, plot, and character comprehension, outperforming the runner-up Qwen2-VL by 5.3\%. However, without subtitles, open-source video models like InternVL2, Qwen2-VL, and MiniCPM-V 2.6 can surpass GPT-4o in \texttt{[Visual]} and \texttt{[Comprehensive]} tasks (\textit{see Appendix}). This indicates that open-source multimodal models with optimized visual processing modules are capable of approaching the visual task performance of commercial models, although a significant gap remains in text comprehension and reading capabilities relative to GPT-4o.

\textbf{MLLMs with LIMITED audio processing capabilities can comprehend audio components.}
While most video MLLMs lack audio input support, our research reveals that audio provides limited benefits in video-audio MLLMs. For instance, VideoLLaMA2.1-AV showed decreased performance when audio is included. In contrast, textual inputs like subtitles help models better capture semantic and contextual cues, significantly improving performance(\textit{see Appendix}). Furthermore, inefficiencies in complex multimodal reasoning tasks highlight the challenges of aligning audio with other modalities and underscore the critical role of high-density textual information in MLLMs.

\textbf{{\ApproachName} enhances the performance of both open-source and commercial MLLMs.}
{\ApproachName} uses plot-event and character-temporal chains to extract fine-grained video segments, enhancing the model's ability to reconstruct complex storylines sequentially. Evaluation results show that {\ApproachName} consistently achieves SOTA performance across all tasks on the narrative-driven {\BenchName}, highlighting its adaptability and scalability across various MLLM architectures. However, a notable gap remains between models and human comprehension of intricate narrative causal reasoning and multi-character plot analysis. These findings indicate substantial potential for improving models in capturing detailed event dependencies and multi-character interactions within narrative contexts.

% TODO: 最好能够画一个雷达图，展示所有模型的整体表现
\subsection{Analysis}
To comprehensively examine the factors influencing the series understanding of \textit{Video}-MLLMs in narrative-driven tasks, we design a set of research questions (RQs) for empirical analysis, including:
(1) investigating the consistency of MLLMs in understanding multi-episode series \textbf{(RQ1)};
(2) analyzing the impact of different input modalities on series understanding \textbf{(RQ2)}; and
(3) conducting an ablation study of Dual-Chains in {\ApproachName} \textbf{(RQ3)}.

\textbf{RQ1: Can MLLMs consistently understand multi-episode series?}
We investigate the performance of different models on {\BenchName} across varying episode contexts and report results for a subset of tasks related to multi-episode series, including \textit{[Plot]}, \textit{[Engagement]}, and \textit{[Empathy]}. These tasks require the comprehension of interleaved modalities and the ability to extract relevant information from multiple videos. The increased token count in multi-episode inputs also demands models capable of processing longer context. Tab. \ref{tab:series_results} summarizes the performance, with \textit{detailed results available in the Appendix.}

\begin{table}[!ht] % 使用 table 而不是 table*，并使用 [H] 选项
    \vspace{5pt} % 调整间距，负值表示减少间距
    \centering
    \small
    \renewcommand\tabcolsep{6pt} % column space
    \renewcommand\arraystretch{0.75} % row space
    \resizebox{1.0\linewidth}{!}{ % 适应单栏宽度
        \begin{tabular}{c|c|c|c|c|c}
            \toprule
            \rowcolor{gray!25} 
            \textbf{Model}
            & \textbf{Episodes}
            & \textbf{Plot}
            & \textbf{Eng.}
            & \textbf{Emp.}
            & \textbf{Total} \\
            \midrule
            \multirow{5}{*}{InternVL2}
            & - & 63.1 & 67.5 & 56.8 & 64.0 \\
            & $\text{Prev}_2$ & 60.9 & 69.2 & 54.1 & 63.1 \\
            & $\text{Prev}_1$ & 64.9 & 68.0 & 54.1 & 65.2 \\
            & $\text{Next}_1$ & 61.2 & 66.9 & 67.6 & 63.5 \\
            & $\text{Next}_2$ & 61.2 & 67.5 & 56.8 & 62.9 \\

            \midrule
            \multirow{5}{*}{Qwen2-VL}
            & - & 61.5 & 70.4 & 56.8 & 64.0 \\
            & $\text{Prev}_2$ & 60.9 & 70.4 & 70.3 & 64.6 \\
            & $\text{Prev}_1$ & 62.5 & 70.4 & 64.9 & 65.2 \\
            & $\text{Next}_1$ & 58.2 & 68.0 & 54.1 & 61.0 \\
            & $\text{Next}_2$ & 60.3 & 69.2 & 56.8 & 62.9 \\
            
            \midrule
            \multirow{5}{*}{MiniCPM-V 2.6}
            & - & 64.6 & 66.9 & 45.9 & 64.0 \\
            & $\text{Prev}_2$ & 57.5 & 64.5 & 54.1 & 59.5 \\
            & $\text{Prev}_1$ & 60.3 & 65.1 & 40.5 & 60.5 \\
            & $\text{Next}_1$ & 56.6 & 55.0 & 51.4 & 55.7 \\
            & $\text{Next}_2$ & 56.6 & 51.5 & 48.6 & 54.4 \\
            
            \bottomrule
        \end{tabular}
    }
    \vspace{-6pt} % 调整间距，负值表示减少间距
    \captionsetup{belowskip=-17pt} % 设置标题和之后段落的距离
    \caption{
        \textbf{Performance of Top-Performing MLLMs on Multi-Episode Series Tasks.}
        Plot, Engagement (\textbf{Eng.}), and Empathy (\textbf{Emp.}), and Total scores.
        $\text{Prev}_i$ and $\text{Next}_i$ represent the inclusion of $i$ episodes before and after the target episode, respectively.
        Evaluation for GPT-4o is not feasible due to API limitations.
    }
    \label{tab:series_results}
\end{table}

\begin{table*}[t!]
    \centering
    \small
    \renewcommand\tabcolsep{8pt} % column space
    \renewcommand\arraystretch{0.80} % row space
    \resizebox{1.0\linewidth}{!}{
        \begin{tabular}{c|l|l|l|l|l|l|l}
            \toprule
            % header of table
            \rowcolor{gray!25} 
            \textbf{Model}
            & \multicolumn{1}{c|}{\textbf{Inputs}}
            & \multicolumn{1}{c|}{\textbf{VS}}
            & \multicolumn{1}{c|}{\textbf{SC}}
            & \multicolumn{1}{c|}{\textbf{AU}}
            & \multicolumn{1}{c|}{\textbf{AG}}
            & \multicolumn{1}{c|}{\textbf{CO}}
            & \multicolumn{1}{c}{\textbf{Overall}} \\
            % & \textbf{BLEU-2(\%)}
            % & \textbf{METEOR(\%)}
            % & \textbf{COSINE(\%)} \\

            \midrule

            \multirow{5}{*}{InternVL2}
            & Q only & 42.5 & 43.2 & 45.7 & 43.5 & 43.6 & \textbf{43.6} \\
            & Q, F & 50.2 (\textit{+7.7}) & 51.3 (\textit{+8.1}) & 52.6 (\textit{+6.9}) & 73.6 (\textit{+30.1}) & 55.1 (\textit{+11.5}) & \textbf{55.2} (\textit{+11.6}) \\
            & Q, S & 51.1 (\textit{+8.6}) & 55.0 (\textit{+11.8}) & 57.9 (\textit{+12.2}) & 74.9 (\textit{+31.4}) & 55.1 (\textit{+11.5}) & \textbf{57.7} (\textit{+14.1}) \\
            & Q, F, S & 52.0 (\textit{+9.5}) & 56.8 (\textit{+13.6}) & 57.7 (\textit{+12.0}) & 78.6 (\textit{+35.1}) & 57.7 (\textit{+14.1}) & \textbf{59.2} (\textit{+15.6}) \\
            & Q, F, S, TC & 55.3 (\textit{+12.8}) & 61.4 (\textit{+18.2}) & 59.9 (\textit{+14.2}) & 76.9 (\textit{+33.4}) & 60.3 (\textit{+16.7}) & \textbf{61.7} (\textit{+18.1}) \\
            
            \midrule
            
            \multirow{5}{*}{Qwen2-VL}
            & Q only & 44.7 & 37.4 & 42.8 & 40.5 & 35.3 & \textbf{41.1} \\
            & Q, F & 52.9 (\textit{+8.2}) & 55.0 (\textit{+17.6}) & 56.4 (\textit{+13.6}) & 72.2 (\textit{+31.7}) & 57.7 (\textit{+22.4}) & \textbf{57.7} (\textit{+16.6}) \\
            & Q, S & 52.2 (\textit{+7.5}) & 54.0 (\textit{+16.6}) & 59.1 (\textit{+16.3}) & 75.9 (\textit{+35.4}) & 53.8 (\textit{+18.5}) & \textbf{58.1} (\textit{+17.0}) \\
            & Q, F, S & 55.7 (\textit{+11.0}) & 57.5 (\textit{+20.1}) & 58.6 (\textit{+15.8}) & 75.3 (\textit{+34.8}) & 59.6 (\textit{+24.3}) & \textbf{60.3} (\textit{+19.2}) \\
            & Q, F, S, TC & 53.5 (\textit{+8.8}) & 61.9 (\textit{+24.5}) & 58.9 (\textit{+16.1}) & 74.2 (\textit{+33.7}) & 63.5 (\textit{+28.2}) & \textbf{60.9} (\textit{+19.8}) \\
            
            \midrule
            
            \multirow{5}{*}{MiniCPM-V 2.6}
            & Q only & 46.5 & 40.9 & 47.0 & 39.5 & 40.4 & \textbf{43.6} \\
            & Q, F & 48.4 (\textit{+1.9}) & 53.1 (\textit{+12.2}) & 54.7 (\textit{+7.7}) & 73.2 (\textit{+33.7}) & 56.4 (\textit{+16.0}) & \textbf{55.6} (\textit{+12.0}) \\
            & Q, S & 48.7 (\textit{+2.2}) & 53.4 (\textit{+12.5}) & 55.5 (\textit{+8.5}) & 72.9 (\textit{+33.4}) & 51.3 (\textit{+10.9}) & \textbf{55.4} (\textit{+11.8}) \\
            & Q, F, S & 53.3 (\textit{+6.8}) & 57.3 (\textit{+16.4}) & 57.7 (\textit{+10.7}) & 76.6 (\textit{+37.1}) & 55.1 (\textit{+14.7}) & \textbf{59.1} (\textit{+15.5}) \\
            & Q, F, S, TC & 50.4 (\textit{+3.9}) & 55.9 (\textit{+15.0}) & 55.7 (\textit{+8.7}) & 74.9 (\textit{+35.4}) & 56.4 (\textit{+16.0}) & \textbf{57.3} (\textit{+13.7}) \\
            
            \midrule
            
            \multirow{5}{*}{GPT-4o}
            & Q only & 46.4 & 44.3 & 46.0 & 38.1 & 41.7 & \textbf{44.1} \\
            & Q, F & 52.4 (\textit{+6.0}) & 56.4 (\textit{+12.1}) & 56.0 (\textit{+10.0}) & 72.2 (\textit{+34.1}) & 47.4 (\textit{+5.7}) & \textbf{56.9} (\textit{+12.8}) \\
            & Q, S & 55.3 (\textit{+8.9}) & 61.2 (\textit{+16.9}) & 61.1 (\textit{+15.1}) & 75.3 (\textit{+37.2}) & 57.1 (\textit{+15.4}) & \textbf{61.3} (\textit{+17.2}) \\
            & Q, F, S & 55.8 (\textit{+9.4}) & 62.8 (\textit{+18.5}) & 60.6 (\textit{+14.6}) & 79.9 (\textit{+41.8}) & 59.6 (\textit{+17.9}) & \textbf{62.8} (\textit{+18.7}) \\
            & Q, F, S, TC & 57.3 (\textit{+10.9}) & 65.1 (\textit{+20.8}) & 65.0 (\textit{+19.0}) & 76.9 (\textit{+38.8}) & 55.1 (\textit{+13.4}) & \textbf{63.8} (\textit{+19.7}) \\
            
            \bottomrule
            
        \end{tabular}
    }
    \vspace{-5pt} % 调整间距，负值表示减少间距
    % 这里给出了总表中效果比较好的几个模型的结果，包含InternVL2, Qwen2-VL, MiniCPM-V 2.6, GPT-4o
    \caption{
        \textbf{Performance of 4 Top-Performing MLLMs with Different Input Modalities.}
        (a) Question text (\textbf{Q}) only,
        (b) \textbf{Q} with Frames (\textbf{F}),
        (c) \textbf{Q} with Subtitles (\textbf{S}),
        (d) \textbf{Q} with \textbf{F} and \textbf{S},
        (e) \textbf{Q} with \textbf{F}, \textbf{S}, and Thematic-Character information (\textbf{TC}).
    }
    \vspace{-0.5cm} % 调整间距，负值表示减少间距

    \label{tab:refer_results}
\end{table*}

Experimental results show improved performance when earlier episodes are used as input. We attribute this improvement to the richer contextual information in earlier episodes, which enhances comprehension of plot-related tasks. In contrast, later episodes, which are not available in the current context, may introduce irrelevant details and hinder comprehension. Notably, Qwen2-VL is the only model that shows improved performance with earlier episodes, demonstrating its ability to perform cross-context reasoning, while InternVL2 and MiniCPM-V 2.6 experience declines under the same conditions. This discrepancy highlights the importance of series-based narrative tasks over individual video understanding tasks for evaluating MLLMs' reasoning capabilities.

\textbf{RQ2: How do different modalities affect series understanding?}
Most benchmarks use only video frames and questions as inputs, often neglecting other crucial information. Similar to Video-MME \cite{fu2024video} and VideoVista \cite{li2024videovista}, we generated subtitle information and obtained thematic and character descriptions during video collection, ensuring synchronization with visual content. We explore how additional descriptive elements influence series understanding by varying input modalities. Experimental results are presented in Tab. \ref{tab:refer_results}. We observed:

\textbf{(1) Video frames or subtitles significantly improve accuracy and consistency.} Supplementing text-only questions with frames (Q, F) or subtitles (Q, S) leads to performance gains across all models, showing the validity of these elements.
\textbf{(2) Textual components outweigh visual information.} Combining text questions with subtitles (Q, S) yields higher accuracy than with frames (Q, F), which is likely due to the narrative nature of videos, where subtitles function as a \textbf{SCREENPLAY} that aids storyline understanding.
\textbf{(3) MLLMs process video information similarly to human cognition.} As humans benefit from subtitles in films, models improve comprehension of thematic and character information. Integrating thematic and character data (Q, F, S, TC) with frames and subtitles (Q, F, S) significantly boosts performance, achieving optimal results across all modalities.
% TODO: 最好能够画一个指代的准确率的图，目前考虑用柱状图

\textbf{RQ3: How do dual-chains impact {\ApproachName}?}
We conduct a comprehensive analysis of the Plot-Event Chain and Character-Temporal Chain to assess their effectiveness in {\ApproachName} during inference. Ablation results are presented in Tab. \ref{tab:cot_results}.
Both the dual-chains are essential for the reasoning process. Comprehensive series understanding requires precise alignment of events and characters, omitting either chain decreases performance, highlighting the importance of joint events and characters for effective reasoning.
\begin{table}[!ht] % 使用 table 而不是 table*，并使用 [H] 选项
    \vspace{-9pt} % 调整间距，负值表示减少间距
    \centering
    \small
    \renewcommand\tabcolsep{4.0pt} % column space
    \renewcommand\arraystretch{0.75} % row space
    \resizebox{1.0\linewidth}{!}{ % 适应单栏宽度
        \begin{tabular}{l|c|c|c|c|c|c}
            \toprule
            \rowcolor{gray!25} 
            \multicolumn{1}{c|}{\textbf{Model}}
            & \multicolumn{1}{c|}{\textbf{VS}}
            & \multicolumn{1}{c|}{\textbf{SC}}
            & \multicolumn{1}{c|}{\textbf{AU}}
            & \multicolumn{1}{c|}{\textbf{AG}}
            & \multicolumn{1}{c|}{\textbf{CO}}
            & \multicolumn{1}{c}{\textbf{Overall}} \\
            \midrule
            InternVL2\red{{$^\dag$}} & 76.5 & 71.4 & 67.3 & 81.1 & 66.7 & \textbf{73.3} \\
            % {\hskip 1em}(w/o) Retriever & 76.8 & 68.2 & 67.8 & 81.4 & 61.0 & \textbf{72.4} \\
            {\hskip 1em}(w/o) Cha-Temp & 71.1 & 66.3 & 64.1 & 75.4 & 62.4 & \textbf{68.5} \\
            {\hskip 1em}(w/o) Plot-Event & 69.3 & 67.4 & 62.3 & 80.4 & 65.2 & \textbf{68.8} \\
            
            \midrule
            Qwen2-VL\red{{$^\dag$}} & 75.2 & 73.5 & 69.7 & 80.9 & 67.4 & \textbf{73.9} \\
            % {\hskip 1em}(w/o) Retriever & 74.7 & 73.0 & 72.3 & 81.1 & 63.8 & \textbf{73.9} \\
            {\hskip 1em}(w/o) Cha-Temp & 64.4 & 66.0 & 62.0 & 79.6 & 61.0 & \textbf{66.4} \\
            {\hskip 1em}(w/o) Plot-Event & 61.7 & 67.6 & 58.6 & 79.3 & 64.5 & \textbf{65.4} \\
            
            \midrule
            MiniCPM-V 2.6\red{{$^\dag$}} & 76.3 & 68.2 & 66.5 & 82.5 & 60.0 & \textbf{72.0} \\
            % {\hskip 1em}(w/o) Retriever & 70.9 & 69.4 & 63.5 & 80.7 & 60.3 & \textbf{69.7} \\
            {\hskip 1em}(w/o) Cha-Temp & 60.1 & 58.7 & 56.7 & 78.9 & 54.2 & \textbf{61.6} \\
            {\hskip 1em}(w/o) Plot-Event & 57.6 & 58.1 & 57.4 & 80.6 & 56.5 & \textbf{61.3} \\
            
            \midrule
            GPT-4o\red{{$^\dag$}} & 78.6 & 76.1 & 73.8 & 82.1 & 61.7 & \textbf{76.2} \\
            % {\hskip 1em}(w/o) Retriever & 76.2 & 74.4 & 74.1 & 80.6 & 58.9 & \textbf{74.8} \\
            {\hskip 1em}(w/o) Cha-Temp & 70.1 & 66.7 & 68.1 & 76.9 & 58.3 & \textbf{69.0} \\
            {\hskip 1em}(w/o) Plot-Event & 71.1 & 76.5 & 66.3 & 86.5 & 55.8 & \textbf{72.3} \\

            \bottomrule

        \end{tabular}
    }
    \vspace{-5pt} % 调整间距，负值表示减少间距
    \caption{
        \textbf{Ablation study of dual-chains on {\BenchName}.}
        \red{{$\dag$}}: MLLMs using the PC-DCoT framework during inference.
    }
    \vspace{-0.5cm} % 调整间距，负值表示减少间距
    \label{tab:cot_results}
\end{table}

\section{Conclusion}
\label{sec:conclusion}
To address the challenge of narrative understanding across multi-video series, this paper introduces \textbf{\BenchName}, a novel benchmark designed to comprehensively evaluate MLLMs' understanding of narrative structures and character relationships across series. We further propose \textbf{\ApproachName}, an adaptable and robust framework that enables MLLMs to achieve over a 10\% improvement on {\BenchName}.
Experiments across a wide range of video-MLLMs reveal significant challenges in comprehending narrative structures and character relationships within series, emphasizing the necessity for continued advancements in this area.
We hope that our {\BenchName} and {\ApproachName} will inspire future research focused on enhancing the capacity of MLLMs to understand narrative-driven drama series.

\section*{Acknowledgements}
This work is supported by “Pioneer” and “Leading Goose” R\&D Program of Zhejiang (No. 2024C01020), the National Natural Science Foundation of China (No. 62406015), research funding from Kuaishou Technology, the Emergency Management Research and Development Project of Zhejiang Province (No. 2024YJ018).

{
    \small
    \bibliographystyle{ieeenat_fullname}
    \bibliography{main}
}

\appendix
\clearpage
\section{More Details of \BenchName}
\subsection{Task Dimension Definitions}

Modern videos have diverse and intricate elements, including visuals, scripts, audio, and post-production enhancements. To facilitate a more comprehensive evaluation of large models that aligns with the diverse modalities present in contemporary videos~\cite{FULWILER201239,mak2012visual}, we categorize the tasks into five major dimensions: Visuals, Script, Audio, Augmentation, and Comprehension.

\textbf{1. Visuals.} Visuals focus on understanding and analyzing visual content in the video. \textbf{1.1 Figures}: The model needs to analyze the figures appearing in the video:  (1.1.1) Actions, analyzing what the figures are doing; (1.1.2) Interactions, understanding interactions between figures. \textbf{1.2 Scenes}: The model needs to recognize changes in scenes and spatiotemporal shifts: (1.2.1) Scene transitions, distinguishing between different scene changes; (1.2.2) Spatiotemporal shifts, understanding changes in time and space. \textbf{1.3 Objects}: The model should recognize and track the state of objects in the video: (1.3.1) Presence, confirming whether certain objects are present; (1.3.2) Interaction, analyzing how objects interact with characters.

\textbf{2. Script.} Script assesses the model's understanding of the background, plot, and characters within the video. \textbf{2.1 Background}: The model needs to understand the setting of the story: (2.1.1) World-building, analyzing the overarching background framework constructed in the video; (2.1.2) Time and location, determining the time and place of the story. \textbf{2.2 Plot}: The model needs to analyze the development and complexity of the story: (2.2.1) Plot development, understanding how the story unfolds; (2.2.2) Foreshadowing and payoff, identifying and understanding the setup and resolution of plot elements; (2.2.3) Twists and conflicts, analyzing key turning points and conflicts in the story; (2.2.4) Climaxes and build-ups, identifying the climax and its preceding buildup; (2.2.5) Suspense and continuity, analyzing the suspense and how different scenes transition; (2.2.6) Emotional dynamics, recognizing emotional peaks in the story. \textbf{2.3 Characters}: The model needs to analyze the characters in the story: (2.3.1) Reference, identifying characters in context; (2.3.2) Motivations, analyzing the reasons behind characters’ actions.

\textbf{3. Audio.} Audio elements assess the model’s understanding of sound-based information in the video, including dialogues, background music, and sound effects. \textbf{3.1 Dialogue}: The model needs to understand and analyze the details of dialogues: (3.1.1) Dialogue Attribution, matching dialogue with the character speaking; (3.1.2) Pronoun references, understanding the pronouns or references used in conversations; (3.1.3) Tone and emotion, analyzing the tone and emotional shifts in dialogue. \textbf{3.2 Music}: The model needs to analyze the role of music in the video: (3.2.1) Atmosphere, assessing how background music influences the emotional atmosphere of the scene. \textbf{3.3 Sound Effects}: The model should understand the role of sound effects: (3.3.1) Impact, analyzing how sound effects enhance a scene's emotional or narrative aspects.

\textbf{4. Augmentation.} Modern videos are no longer purely composed of footage; many post-production elements, such as subtitles, labels, and special effects, are added. The model needs to understand and utilize this information to enhance video comprehension.  \textbf{4.1 Subtitles}: The model needs to handle subtitles: (4.1.1) Recognition, accurately recognizing and understanding subtitle information. \textbf{4.2 Labels}: The model needs to understand the annotations in the video: (4.2.1) Purpose, analyzing how labels convey specific information. \textbf{4.3 VFX}: The model should analyze the use of special effects: (4.3.1) Effectiveness, evaluating how special effects influence the visual experience.

\textbf{5. Comprehension.} Comprehension tasks evaluate the model’s overall grasp of the video. \textbf{5.1 Engagement}: The model needs to infer viewers' interest in the plot's development: (5.1.1) Future predictions, predicting future developments based on current information; (5.1.2) Current interpretation, assessing the model’s understanding of the current storyline. \textbf{5.2 Empathy}: The model needs to understand the emotional connection between the audience and the characters: (5.2.1) Character resonance, analyzing the emotional state of characters and generating empathy.

\begin{figure}[t]
    \centering
    % \vspace{-0.3cm}
    \includegraphics[width=1.0\linewidth
    ]{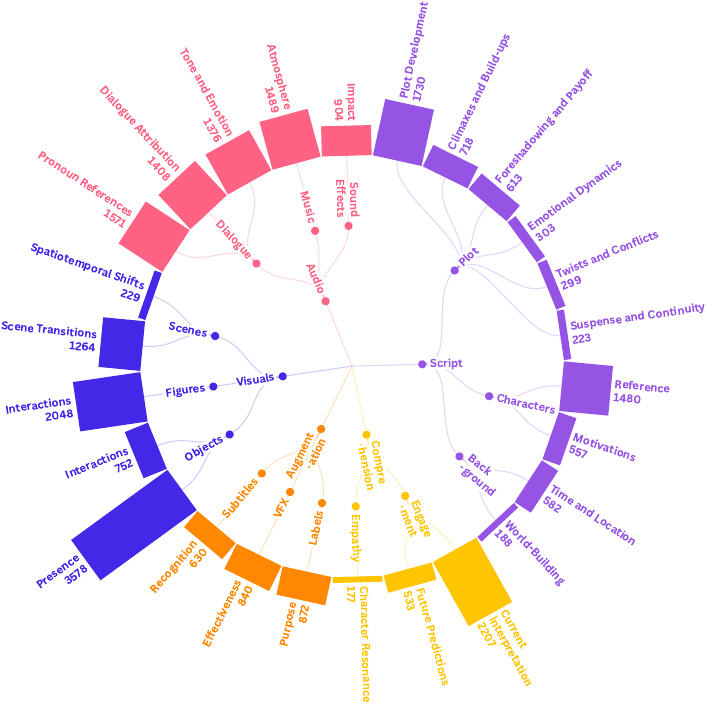}
    \vspace{-0.7cm}
    \caption{
    \textbf{Task Dimension in \BenchName.}
    Detailed sample count for each task in \BenchName.
    }
    \label{fig:Task_Dimension}
    \vspace{-0.5cm}
\end{figure}

\subsection{Data description}
To obtain high-quality series data, we sourced series-format videos from the professional video platform, \textit{Kuaishou}\footnote{https://www.kuaishou.cn}. To ensure higher-quality series, we selected content only from series created by authors with a substantial follower base, as well as videos meeting specific standards for likes and comments. Specifically, we only included series created by authors with more than 10 million followers, where each episode within the series maintained a minimum of 5,000 likes and at least 1,000 comments. 

\begin{figure}[t]
    \centering
    % \vspace{-0.3cm}
    \includegraphics[width=1.0\linewidth
    ]{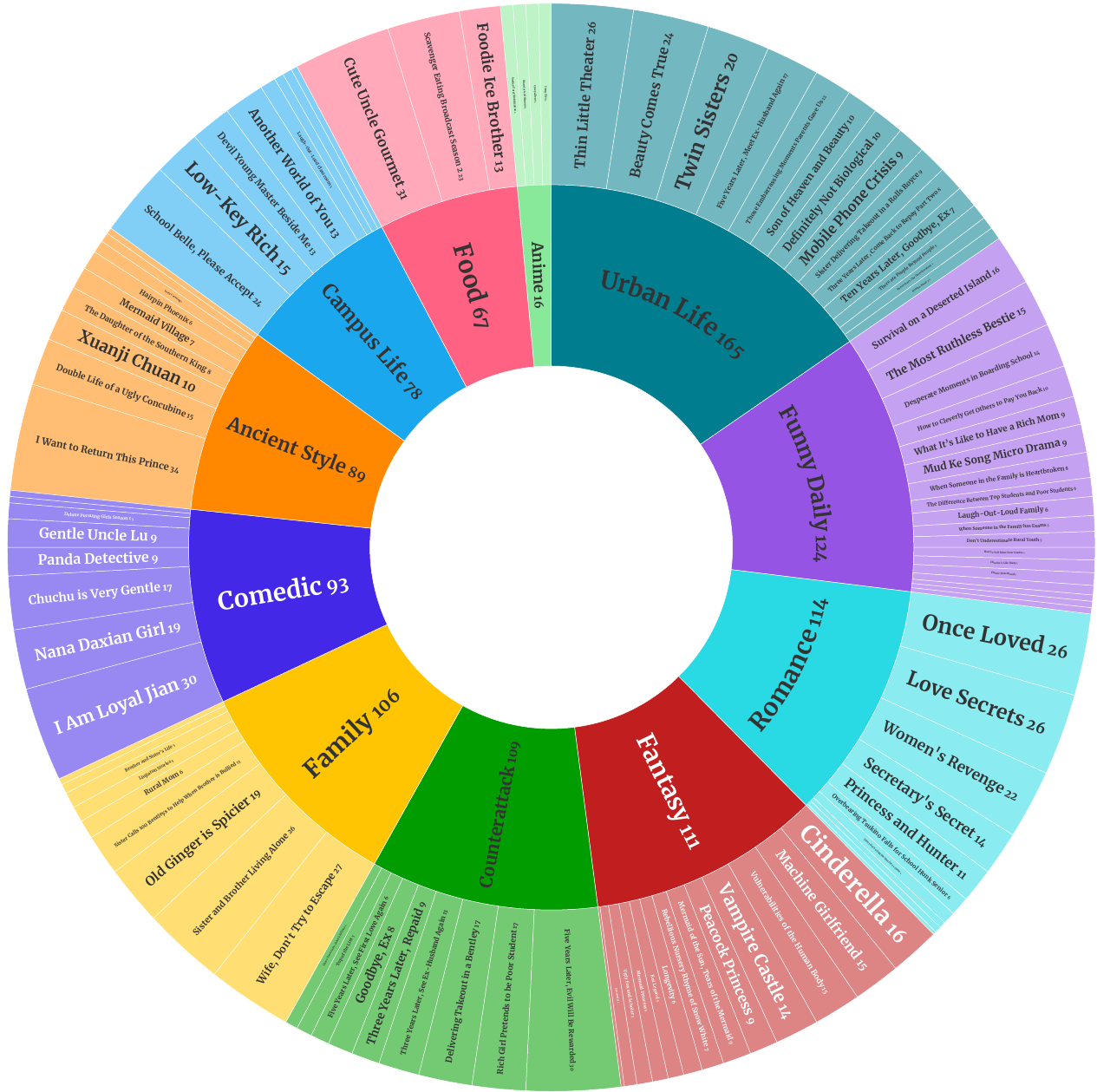}
    \vspace{-0.7cm}
    \caption{
    \textbf{Video Categories in \BenchName.}
    The videos in \BenchName\ are categorized into two main types: thematic videos and series videos. These encompass 11 of the most popular video themes, including Urban Life, Romance, Fantasy, Counterattack, Family, Ancient Style, Campus Life, Anime, Funny Daily, Short Drama, and Food. Each series is accompanied by a number indicating the total count of videos within that series.
    }
    \label{fig:tube}
    \vspace{-0.5cm}
\end{figure}

To further ensure a balanced representation across genres, we selected series relatively evenly from the top 11 most popular genres. Additionally, we controlled the number of episodes within each series, ranging from 2 to 50 episodes per series. This approach provided a diverse yet structured dataset. Ultimately, we curated a total of 105 series, encompassing 1,072 videos. For more detailed insights into the distribution, refer to Fig. \ref{fig:tube}.

\section{More Details of Data Annotation}
To ensure the quality of SeriesBench, we invited many professional data annotators to participate in the annotation process for SeriesBench. The process began with comprehensive training sessions, during which annotators were introduced to the task objectives and detailed annotation guidelines. Following this, a trial annotation phase was conducted to evaluate the annotators' performance and select those most suited to the task. Based on the issues identified during the trial, additional training or personalized guidance was provided to address specific challenges and improve annotation quality. After these iterative steps, we finalized a team of 32 top-performing annotators who demonstrated the highest annotation quality for the main labeling task. Upon completion of the annotation, we conducted a rigorous quality inspection, during which annotations that failed to meet the required standards were either sent back for re-annotation or discarded entirely. Finally, a sampling test of the dataset showed that 96\% of the annotations met our stringent requirements, confirming the high quality of the labeled data in SeriesBench.

\subsection{Annotation Interface}
\begin{figure*}[thp]
    \centering
    % \vspace{-0.3cm}
    \includegraphics[width=0.87\textwidth]{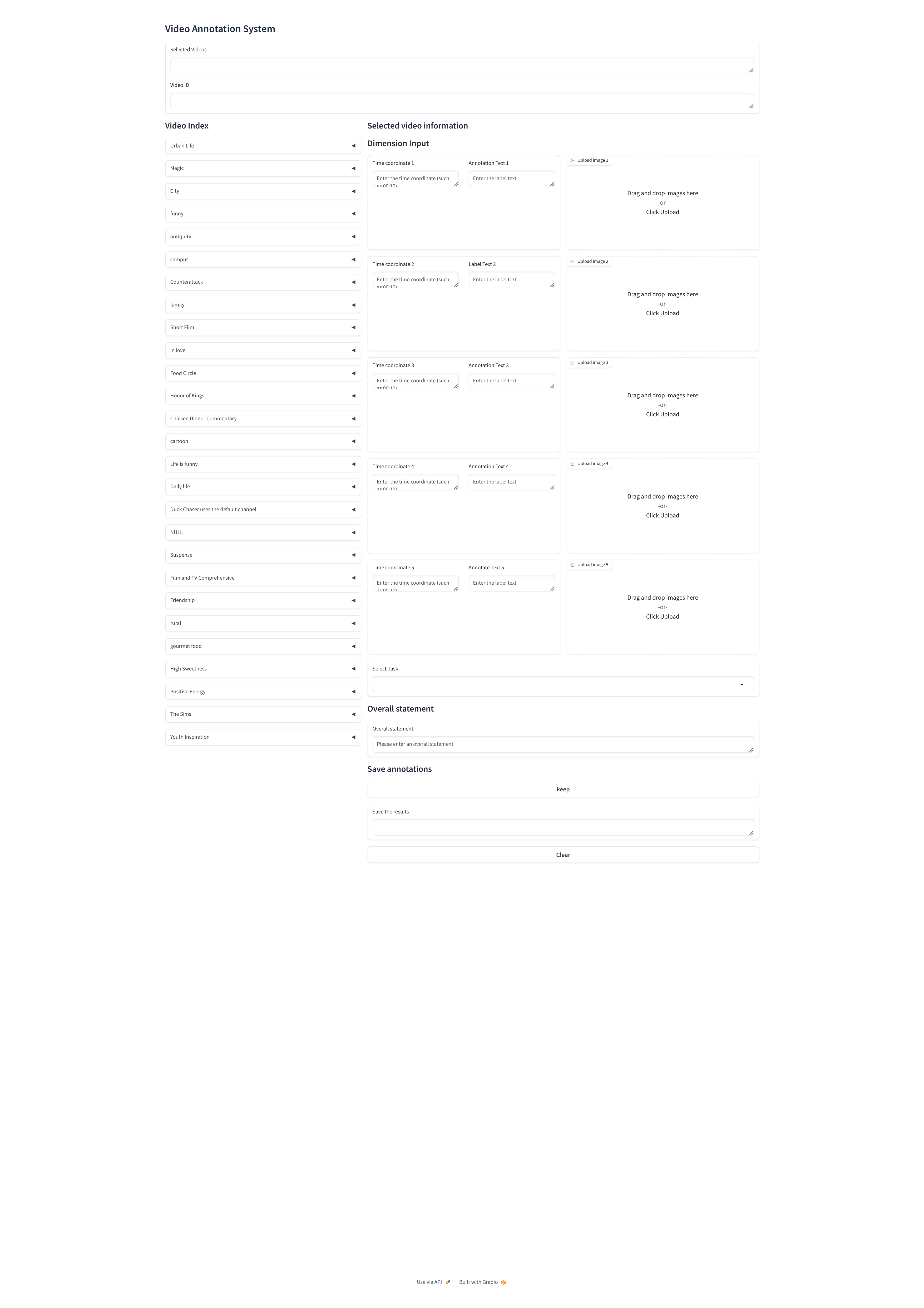}
    \vspace{-0.3cm}
    \caption{
    \textbf{ Illustration of the web-based annotation interface.} 
   Displaying parsed video information for reference, a hierarchical video index for navigation, and an annotation area for detailed content analysis and categorization.
    }
    \label{fig:gradio}
    \vspace{-0.3cm}
\end{figure*}

To facilitate annotation and management, we designed a dedicated annotation interface, as illustrated in Fig. \ref{fig:gradio}. At the top of the interface, information extracted from the selected video is displayed, offering annotators a convenient reference for comparison and ensuring accurate labeling. On the left side, a hierarchical video index organizes content by genre at the primary level, enabling annotators to efficiently navigate and select video series within each category. The right side serves as the annotation workspace, where videos are meticulously analyzed and annotated across multiple dimensions. Each annotation entry includes precise timestamps, detailed descriptions of scenes or characters, and associated character portrait screenshots to ensure comprehensive labeling. After completing these elements, annotators select the most appropriate task category and craft a concise declarative statement sentence that integrates and contextualizes the annotated content, resulting in a cohesive and informative annotation output. Once the process is complete, annotators can save their work by clicking the ``keep'' button or reset the interface for the next task using the ``clear'' button, streamlining the workflow for subsequent annotations.

\subsection{Manual annotation}
\begin{figure*}[thp]
    \centering
    % \vspace{-0.3cm}
    \includegraphics[width=0.87\textwidth]{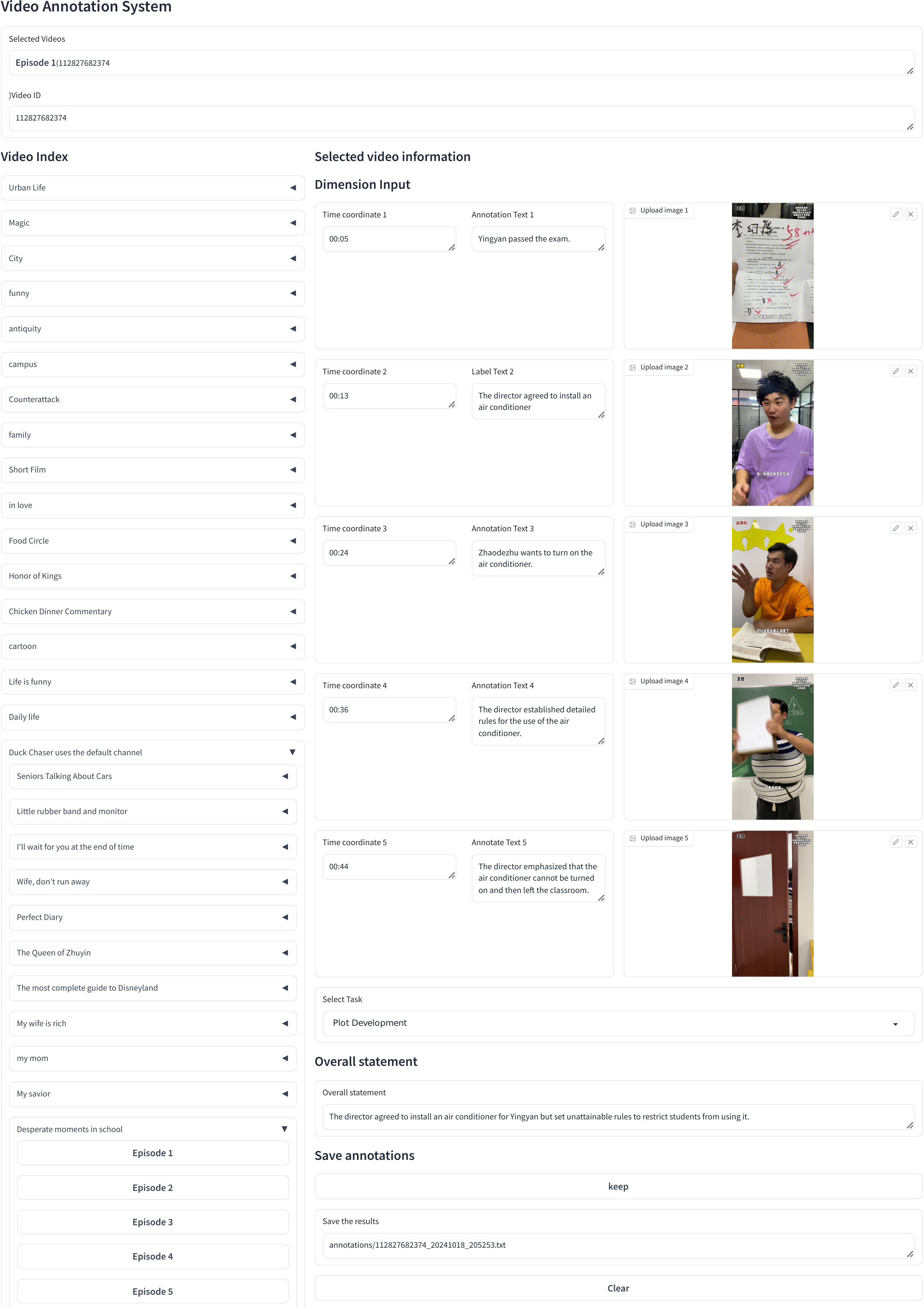}
    \vspace{-0.3cm}
    \caption{
    \textbf{Example of the annotation process.} 
    Showing the selection of a video series, sequential annotation of episodes, and detailed documentation of key events, timelines, and character interactions, culminating in task categorization and a synthesized declarative summary to enhance content representation.
    }
    \label{fig:gradio_annotation}
    \vspace{-0.3cm}
\end{figure*}

As illustrated in Fig. \ref{fig:gradio_annotation}, this example demonstrates a specific annotation process in detail. Once the annotator selects a video series, the interface expands to display all episodes within that series. The annotator is required to begin with the first episode and progress sequentially through to the final one, ensuring comprehensive coverage of the entire series.  When the button for a particular episode is clicked, the interface redirects to the corresponding video playback page, where the annotator must watch the entire video before initiating the annotation process. Guided by the established annotation guidelines, the annotator meticulously documents critical aspects of the video, including timelines, character interactions, and major events. This process also involves capturing relevant screenshots of key characters to provide visual references that enrich the annotation.

After completing the detailed annotations, the annotator selects the most suitable task category that aligns with the content. In this example, ``Plot development'' was chosen, reflecting the emphasis on narrative progression and storyline evolution. As a final step, the annotator crafts a concise declarative statement sentence that synthesizes all the annotated content, ensuring that it reflects a well-connected summary and contains multi-step inferential insights. This approach guarantees that the resulting statement offers a high-information, comprehensive representation of the video content, enhancing the depth and quality of the benchmark annotations.

\section{More Details of Heuristic Baselines}
We provide further details on the heuristic baselines introduced in Section 5.2, inspired by MathVista~\cite{lu2023mathvista}: \textit{Random Choice}, \textit{Frequent Guess}, and \textit{Human Evaluation}. These baselines are critical for comparing model performance on the {\BenchName} tasks.

\textbf{Random Choice.} The Random Choice baseline selects an answer randomly from the answer pool for each question and averages the results over five trials. This baseline serves as a simple reference for evaluating model performance, representing the expected outcome of random guessing.

\textbf{Frequent Guess.} Based on the option distribution in each task category of the {\BenchName} training set, we select the most frequently occurring option as the predicted answer for the corresponding task in the test set. This baseline demonstrates whether the option distribution in {\BenchName} is balanced and serves as a straightforward yet informative reference for evaluating model performance, representing the expected outcome of consistently selecting the most common answer.

\begin{figure*}[thp]
  \centering
  % \vspace{-0.3cm}
  \includegraphics[width=0.95\textwidth]{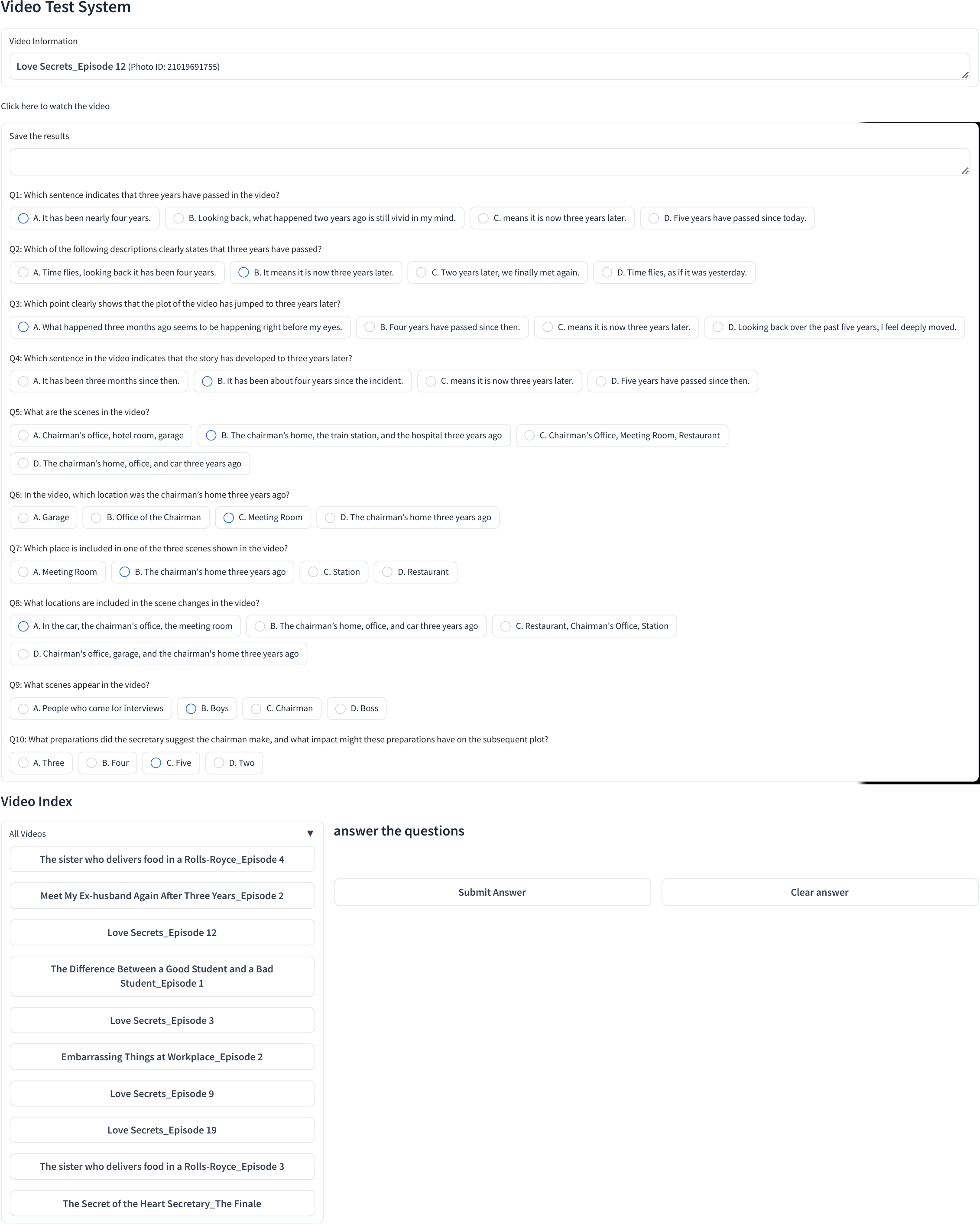}
  \vspace{-0.3cm}
  \caption{
  \textbf{The manual evaluation interface.} 
   Comparing model results with human performance, displaying randomly selected videos, corresponding questions, and indexed video links for evaluators to review and submit their responses.
  }
  \label{fig:gradio_human}
  \vspace{-0.3cm}
\end{figure*}

\textbf{Human Evaluation.} The human evaluation baseline reflects human performance on {\BenchName}, serving as a reliable upper bound for assessing model capabilities. To facilitate this process, we developed a structured manual evaluation workflow with a user-friendly interface, illustrated in Fig.~\ref{fig:gradio_human}.
Each evaluation session involves a random sampling of 10 videos from the test set. Upon selecting a video, the evaluator is presented with the corresponding question and a link to the indexed video. After reviewing the video, the evaluator submits their answer with a single click and completes the assessment. This streamlined workflow ensures efficiency and consistency, providing a robust basis for direct comparison with model-generated results.

\section{More Details of Evaluation Settings}

\subsection{Prompt for Evaluation}
\begin{figure*}[thp]
    \centering
    % \vspace{-0.3cm}
    \includegraphics[width=0.95\textwidth]{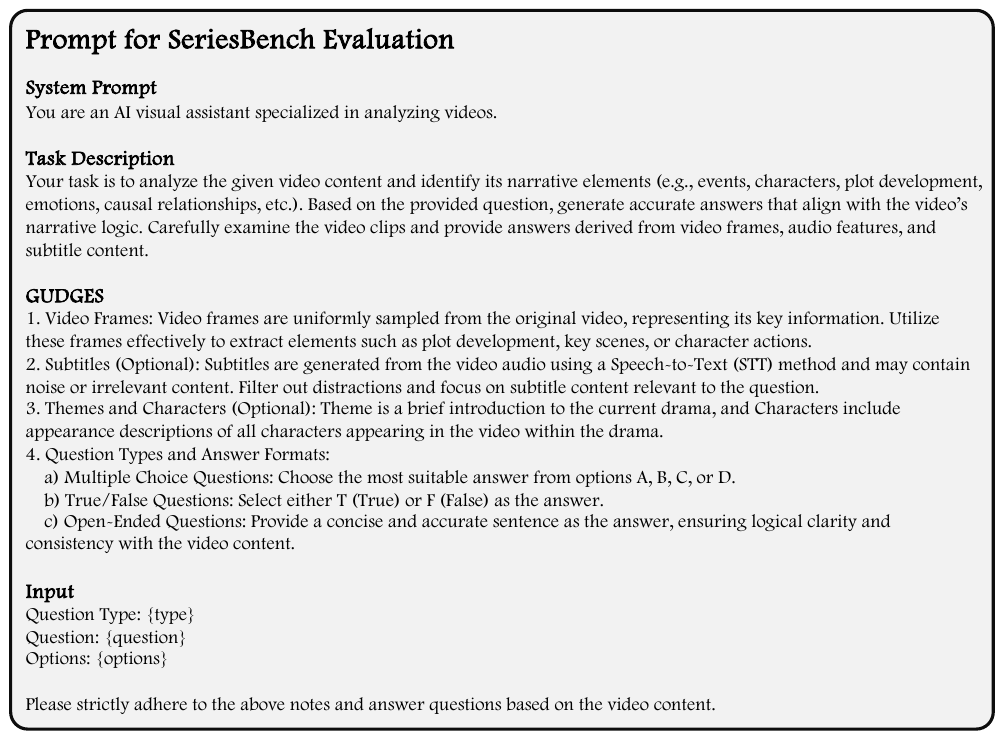}
    \vspace{-0.3cm}
    \caption{
        \textbf{The prompt template adopted for Evaluation on {\BenchName}.}
    }
    \label{fig:prompt_eval}
    \vspace{-0.5cm}
\end{figure*}

\begin{figure*}[thp]
  \centering
  % \vspace{-0.3cm}
  \includegraphics[width=0.95\textwidth]{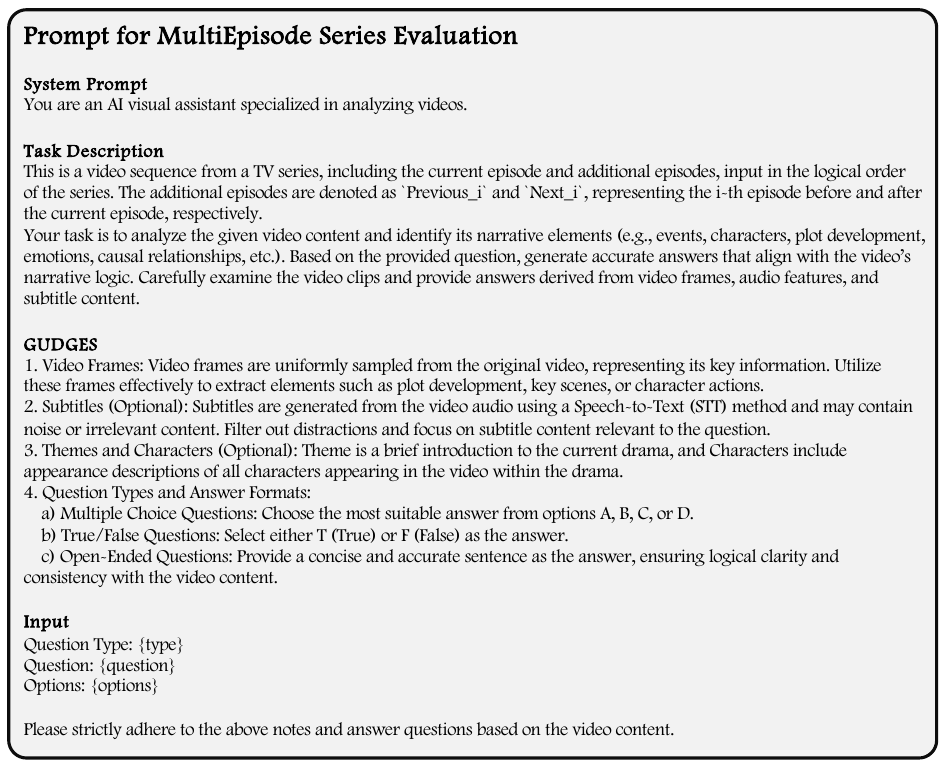}
  \vspace{-0.3cm}
  \caption{
  \textbf{The prompt template adopted for Multi-Episode series tasks.} 
  }
  \label{fig:prompt_multivideo}
  \vspace{-0.5cm}
\end{figure*}

We provide the detailed prompt template for evaluating the model's performance on {\BenchName} in Fig.~\ref{fig:prompt_eval}. Additionally, we provide a prompt template for evaluating the model's performance on multi-video tasks in Fig.~\ref{fig:prompt_multivideo}, tailored to assess the model's ability to analyze and synthesize information across multi-episode series. 
The model input is divided into five parts: \texttt{<frames>}, \texttt{<subtitles>}, \texttt{<theme-chara>}, \texttt{<prompt>}, and \texttt{<question>}. The \texttt{<frames>} part contains the video frames, the \texttt{<subtitles>} part includes the video subtitles, the \texttt{<theme-chara>} part involves the theme and characters of the video, the \texttt{<prompt>} part provides the instruction for the model, and the \texttt{<question>} part consists of the question for the model to answer.
The guidelines for \texttt{<subtitles>} and \texttt{<theme-chara>} in Fig.~\ref{fig:prompt_eval} and Fig.~\ref{fig:prompt_multivideo} are optional and can be adjusted to match the input.

\subsection{Model Inference Settings}

\textbf{GPT-4o and GPT-4o-mini} 
Due to API limitations, we uniformly sampled 50 frames from each video for evaluation on {\BenchName}, testing under both ``without subtitles'' and ``with subtitles'' settings. The model input adopts the format of ``\texttt{<frames> + <subtitles>(optional) + <theme-chara>(optional) + <prompt> + <question>}''.
\\
\textbf{VideoLLaMA2.1-AV}
We investigated how audio influences model performance by extracting the audio tracks from the videos and incorporating them into the model's input. The model input adopts the format of ``\texttt{<frames> + <subtitles>(optional) + <audio>(optional) + <prompt> + <question>}''.
\\
\textbf{Other Open-Source \textit{Video}-MLLMs} 
We adhere to the official inference strategies of these MLLMs. To ensure stable inference for Qwen2-VL, where we observed occasional CUDA-OOM errors under certain conditions, we set the number of input frames to 64 instead of the default setting of 1fps. The model input adopts the format of ``\texttt{<frames> + <subtitles>(optional) + <theme-chara>(optional) + <prompt> + <question>}''.

\subsection{Evaluation Metrics}
\textbf{MultiChoice and Judgment Type Tasks.}
All tasks are formulated as multiple-choice questions, and we adopt accuracy as the primary evaluation metric.
\\
\textbf{Open-Ended Type Tasks.}
For open-ended questions, where the model's responses are required to be concise sentences containing the correct answer, we adopt BLEU-2~\cite{papineni2002bleu}, METEOR~\cite{banerjee2005meteor}, and BERTScore F1~\cite{bert-score} as evaluation metrics to assess model outputs from both lexical and semantic perspectives~\cite{li2024mateval,llm_eval_metrics,llm_monitoring}.
\textbf{BLEU-2} evaluates the bi-gram (two-word phrase) overlap between generated and reference texts, reflecting the degree of lexical similarity and fluency.
\textbf{METEOR}, widely used in machine translation and text generation, considers word-level matches, including stemming and synonym matching, providing a more nuanced and comprehensive evaluation.
\textbf{BERTScore F1} measures semantic similarity by calculating the cosine similarity between the embeddings of the generated and reference sentences in the BERT encoding space, offering an accurate assessment of sentence-level semantic alignment.

\section{Implement of Method}
\subsection{Retriever module of PC-DCoT}

We developed a Retriever module to extract event- and character-related timelines from video frames, aiming to better understand the narrative structure in series. Specifically, leveraging the manually annotated event and character timelines in the training set of {\BenchName}, we selected the corresponding video frames to construct a dataset consisting of 6,046 image-text pairs. To better adapt to Chinese annotations, we refined the module using CN-CLIP{$_\text{ViT-H/14}$}~\cite{chinese-clip} as the base model and finetuned it on the constructed dataset. The training hyperparameters are detailed in Tab.~\ref{tab:clip_hyperparas}.

\begin{table}[!ht] % 使用 table 而不是 table*，并使用 [H] 选项
    \centering
    \small
    \renewcommand\tabcolsep{20pt} % column space
    \renewcommand\arraystretch{0.75} % row space
    \resizebox{0.75\linewidth}{!}{ % 适应单栏宽度
        \begin{tabular}{l|c}
            \toprule
            \rowcolor{gray!25} 
            \textbf\textbf{Hyperparameter} & \textbf{config} \\
            \midrule
            context length & 52 \\
            warmup steps & 100 \\
            batch size & 32 \\
            learning rate & 5e-5 \\
            weight decay & 0.001 \\
            training epochs & 5 \\
            \bottomrule
        \end{tabular}
    }
    \vspace{-6pt} % 调整间距，负值表示减少间距
    \captionsetup{belowskip=-17pt} % 设置标题和之后段落的距离
    \caption{
        \textbf{Hyperparameters for CN-CLIP{$_\text{ViT-H/14}$} finetuning.}
    }
    \label{tab:clip_hyperparas}
\end{table}

\subsection{Mathematical representation of PC-DCoT}

The input to our task consists of two primary components: a video sequence and a set of questions. The video is represented as a collection of frames denoted by

\begin{equation}
  V = \{f_1, f_2, \ldots, f_k\}
  \label{eq:video_input}
\end{equation}

where each element \( f_i \) corresponds to an individual frame within the video sequence. Additionally, the input includes a set of questions \( Q \) that are intended to guide the analysis and understanding of the video content.

For narrative-driven videos, understanding the plot and characters is essential, as they serve as the key drivers of the storyline. Our method begins by utilizing the Multimodal Large Language Model (MLLM) to process raw video frame input \( V = \{f_1, f_2, \ldots, f_k\} \) alongside relevant questions \( Q \). The MLLM's role is to extract and identify key narrative elements, specifically focusing on significant events that are pertinent to the questions and key characters that appear within the narrative context. This extraction process results in two distinct sets: a set of events and a set of characters, both represented as text-based descriptions, represented as

\begin{equation}
  \{\mathcal{E}_1, \mathcal{E}_2, \ldots, \mathcal{E}_m\}, \{\mathcal{C}_1, \mathcal{C}_2, \ldots, \mathcal{C}_n\} = \text{MLLM}_{\text{extract}}(Q, V)
  \label{eq:extraction}
\end{equation}

where \(\{\mathcal{E}_1, \mathcal{E}_2, \ldots, \mathcal{E}_m\}\) denotes the extracted events and \(\{\mathcal{C}_1, \mathcal{C}_2, \ldots, \mathcal{C}_n\}\) denotes the identified characters. 

In narrative-driven videos, character appearances often exhibit discontinuity, whereas events typically unfold through cohesive and interconnected sequences. To align video content with task-relevant events and characters, our approach leverages a video clip model to identify and isolate key scenes. When processing raw video input in conjunction with extracted events related to the given questions, the model searches for frames corresponding to each identified event, represented by

\begin{equation}
  \mathcal{F}_{e_j} = \{ f_i \in V \mid \text{CLIP}_{\text{event}}(f_i, \mathcal{E}_j) \geq \theta_e \}
  \label{eq:event_frame_search}
\end{equation}

where \(\text{CLIP}_{\text{event}}\) measures the relevance between each frame \(f_i\) and event \(\mathcal{E}_j\) using a defined threshold \(\theta_e\). To maintain temporal coherence, these frames are combined with neighboring frames within a temporal window \(\delta\), forming cohesive sets of frames for each event, represented by

\begin{equation}
  E_j = \{ f_{i'} \in V \mid \exists f_i \in \mathcal{F}_{e_j}, |i - i'| \leq \delta \}
  \label{eq:event_aggregation}
\end{equation}

The overall set of event sequences is then expressed as

\begin{equation}
  \mathcal{F}_e = \bigcup_{j=1}^{m} E_j
  \label{eq:event_set}
\end{equation}

In parallel, character tracking is performed by locating and retrieving frames where specific characters appear, based on their portraits. This process identifies frames relevant to each character \(\mathcal{C}_j\), represented by

\begin{equation}
  C_j = \{ f_i \in V \mid \text{CLIP}_{\text{character}}(f_i, \mathcal{C}_j) \geq \theta_c \}
  \label{eq:character_frame_search}
\end{equation}

with \(\text{CLIP}_{\text{character}}\) determining frame relevance to the character using a threshold \(\theta_c\). The aggregated frames for all characters are represented as

\begin{equation}
  \mathcal{F}_c = \bigcup_{j=1}^{n} C_j
  \label{eq:character_set}
\end{equation}

To effectively capture the narrative structure and character dynamics, we construct two distinct chains: the Plot Event Chain of Thought (\(\text{CoT}_E\)) and the Character Temporal Chain of Thought (\(\text{CoT}_C\)).

The \(\text{CoT}_E\) is created by generating detailed descriptions for each event set \(E_j\) derived from the aggregated event frames \(\mathcal{F}_e\). This process is expressed as

\begin{equation*}
\text{CoT}_E = \bigcup_{j=1}^{m} \Big\{ d_j \mid \, d_j = \text{MLLM}_{\text{describe-event}}(E_j), \, E_j \in \mathcal{F}_e, \, 
\end{equation*}   
\begin{equation}  
T_j = [t_j^{\text{start}}, t_j^{\text{end}}], \, \text{where } T_j \text{ denotes the time interval of } E_j \Big\}
    \label{eq:event_cot}
\end{equation}

Here, \(\text{MLLM}_{\text{describe-event}}\) generates a narrative description \(d_j\) for each event \(E_j\), and \(T_j = [t_j^{\text{start}}, t_j^{\text{end}}]\) represents the time interval of the event, capturing its temporal boundaries within the video timeline.

Simultaneously, the \(\text{CoT}_C\) is constructed by describing the behaviors and appearances of each character \(C_k\) based on the frames aggregated in \(\mathcal{F}_c\). This is represented as

\begin{equation*}
  \text{CoT}_C = \bigcup_{k=1}^{n} \Big\{ d_k \mid \, d_k = \text{MLLM}_{\text{describe-character}}(C_k), \,
\end{equation*}
\begin{equation*}
  C_k \in \mathcal{F}_c, \, T_k = \{ t_{k_1}, t_{k_2}, \ldots, t_{k_l} \},
\end{equation*}
\begin{equation}
  \text{where } T_k \text{ is the sequence of frames where } C_k \text{ appears} \Big\}
  \label{eq:character_cot}
\end{equation}

This dual-chain construction captures both the continuous flow of events and the potentially scattered yet significant appearances of characters, providing a comprehensive understanding of narrative progression and character interactions.

To merge the Plot Event Chain of Thought (\(\text{CoT}_E\)) and the Character Temporal Chain of Thought (\(\text{CoT}_C\)) into a unified framework, we utilize the precise temporal intervals associated with events and character appearances. Since each event \(d_j\) in \(\text{CoT}_E\) corresponds to a time interval \([t_j^{\text{start}}, t_j^{\text{end}}]\), it is possible to identify characters \(d_k\) from \(\text{CoT}_C\) whose presence overlaps with this interval, represented by the condition \(T_k \cap [t_j^{\text{start}}, t_j^{\text{end}}] \neq \emptyset\). This temporal alignment enables us to determine which characters interact within the timeframe of each event.

The merging process is defined as:

\begin{equation*}
  \text{PC-DCoT} = \bigcup_{j=1}^{m} \{ d_{j}^{\prime} \mid d_{j}^{\prime} = \text{MLLM}_{\text{aggregate}} ( d_j, 
\end{equation*}
\begin{equation}
  \{ d_k \in \text{CoT}_C \mid T_k \cap [t_j^{\text{start}}, t_j^{\text{end}}] \neq \emptyset \} ), \ d_j \in \text{CoT}_E \}
  \label{eq:pc_dcot}
\end{equation}

Here, \(\text{MLLM}_{\text{aggregate}}\) combines each event \(d_j\) from \(\text{CoT}_E\) with relevant character descriptions \(d_k\) from \(\text{CoT}_C\) that appear within the event's time interval. This aggregation synthesizes the behaviors and occurrences of characters with the narrative context of the event, resulting in a cohesive and enriched representation of the narrative flow and character dynamics.

\subsection{Prompt for {\ApproachName}}

\begin{figure*}[thp]
  \centering
  % \vspace{-0.3cm}
  \includegraphics[width=0.95\textwidth]{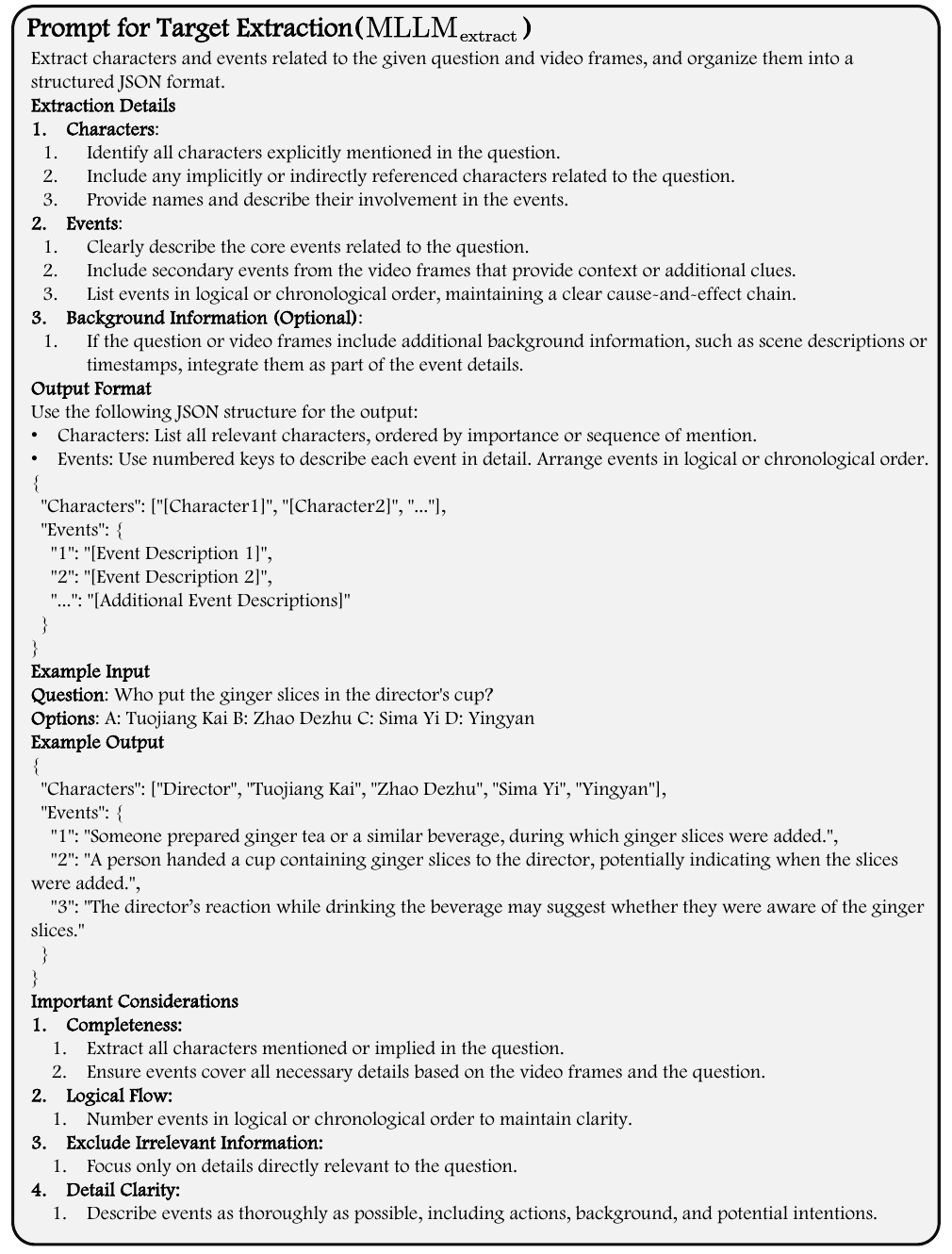}
  \vspace{-0.3cm}
  \caption{
  \textbf{Prompt for Target Extraction.} 
  }
  \label{fig:prompt_pc-dcot_extract}
  \vspace{-0.3cm}
\end{figure*}

\begin{figure*}[thp]
  \centering
  % \vspace{-0.3cm}
  \includegraphics[width=0.95\textwidth]{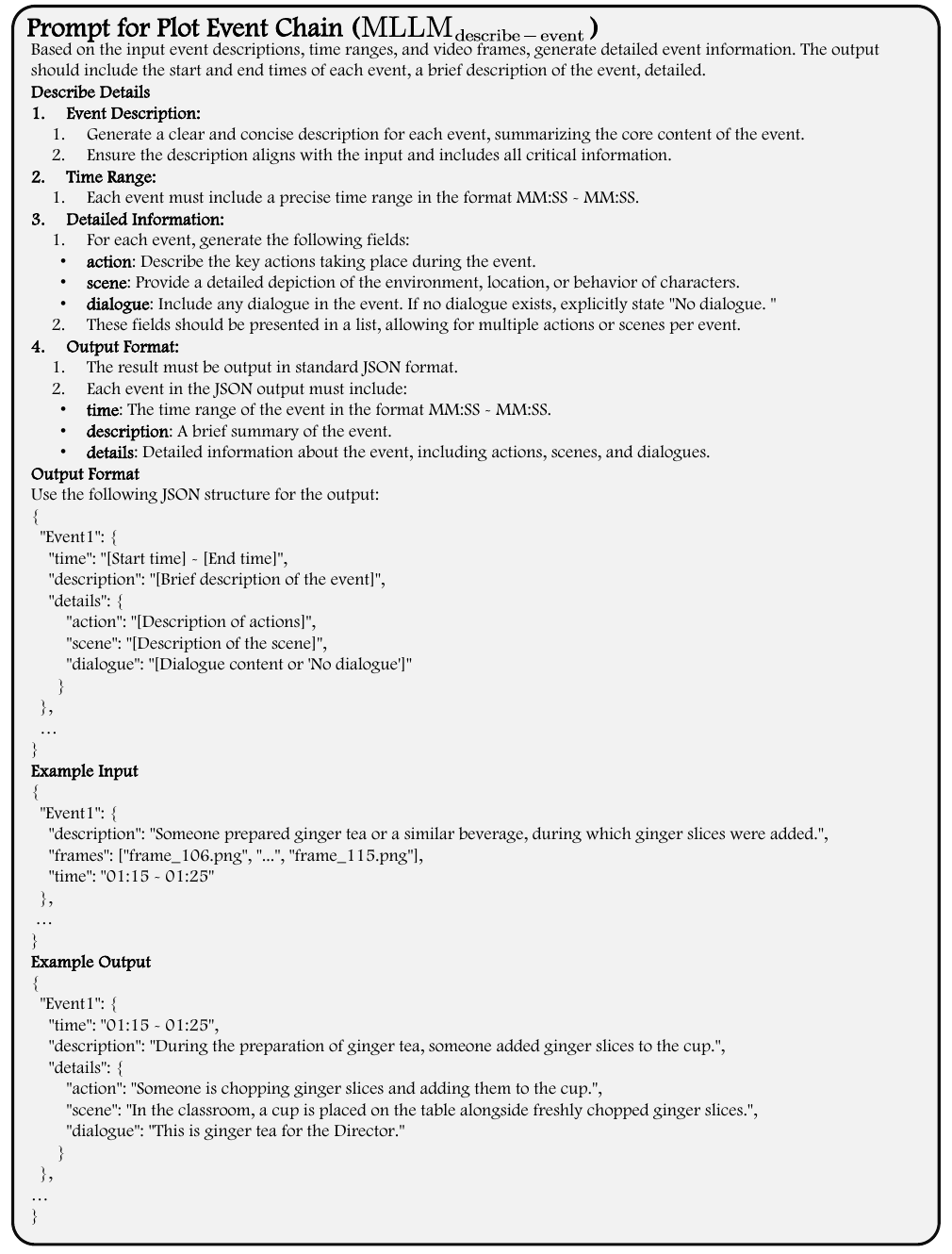}
  \vspace{-0.3cm}
  \caption{
  \textbf{Prompt for Plot Event Chain.} 
    }
  \label{fig:prompt_pc-dcot_describe_event}
  \vspace{-0.3cm}
\end{figure*}

\begin{figure*}[thp]
  \centering
  % \vspace{-0.3cm}
  \includegraphics[width=0.95\textwidth]{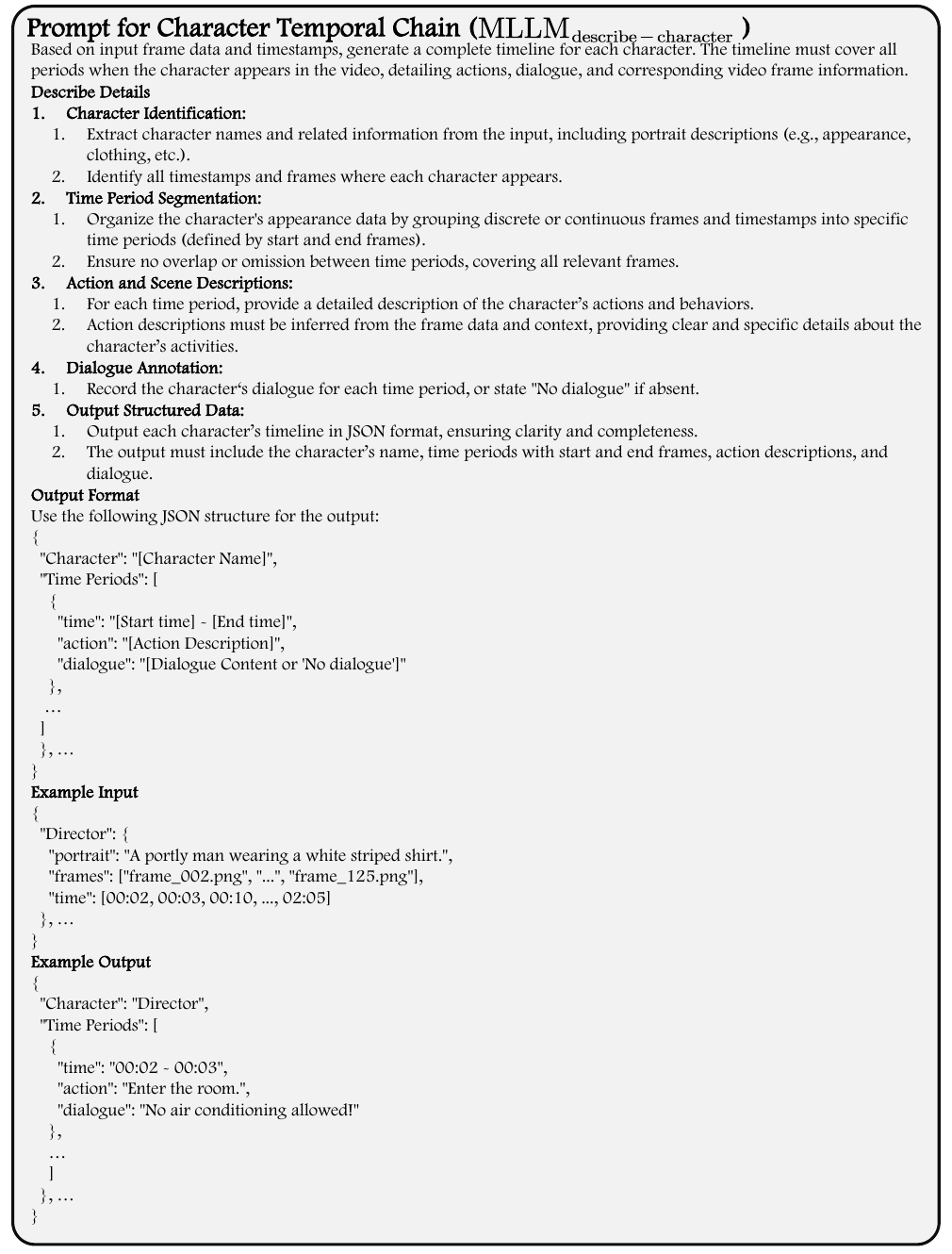}
  \vspace{-0.3cm}
  \caption{
  \textbf{Prompt for Character Temporal Chain.} 
   }
  \label{fig:prompt_pc-dcot_describe_character}
  \vspace{-0.3cm}
\end{figure*}

\begin{figure*}[thp]
  \centering
  % \vspace{-0.3cm}
  \includegraphics[width=0.95\textwidth]{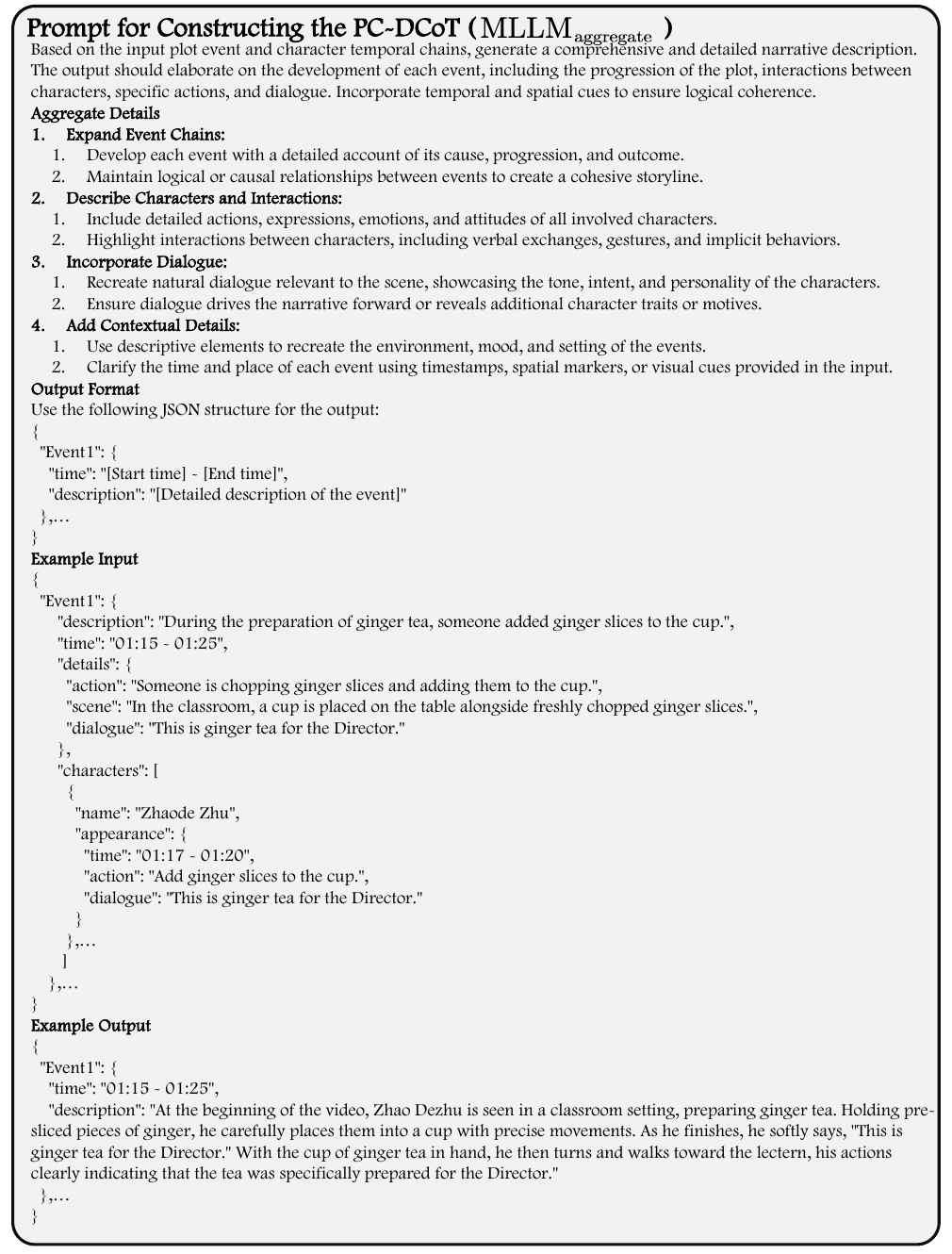}
  \vspace{-0.3cm}
  \caption{
  \textbf{Prompt for Constructing the PC-DCoT.} 
   }
  \label{fig:prompt_pc-dcot_aggregate}
  \vspace{-0.3cm}
\end{figure*}

In the mathematical formulation of PC-DCoT presented above, $\text{MLLM}_{\text{extract}}$, $\text{MLLM}_{\text{describe-event}}$, $\text{MLLM}_{\text{describe-character}}$, and $\text{MLLM}_{\text{aggregate}}$ denote the reasoning processes executed by the MLLM. The corresponding prompts utilized to guide these reasoning processes are depicted in Figs.~\ref{fig:prompt_pc-dcot_extract}, \ref{fig:prompt_pc-dcot_describe_event}, \ref{fig:prompt_pc-dcot_describe_character}, and \ref{fig:prompt_pc-dcot_aggregate}.

\section{More Experimental Results}
In this section, we give more detailed results about the performance of different models on {\BenchName}.

\subsection{Main Results with Subtitles Ablation}
The comprehensive results under the subtitles ablation setting are shown in Tab.~\ref{tab:main_results_sub}. Notably, open-source models, particularly Qwen2-VL, exhibit competitive performance compared to the closed-source model GPT-4o. On the MultiChoice and Judgment tasks of {\BenchName}, Qwen2-VL secures 5 first-place, 5 second-place, and 1 third-place ranking, closely rivaling GPT-4o, which achieves 7 first-place, 3 second-place, and 1 third-place finishes. These results underscore the effectiveness and potential of open-source models in addressing narrative-driven video analysis.

\begin{table*}[!ht]
    \centering
    \vspace{-5pt}
    \small
    \renewcommand\tabcolsep{4pt} % column space
    \renewcommand\arraystretch{0.9} % row space
    \resizebox{1.0\linewidth}{!}{
        \begin{tabular}{l|c|c|cc|cc|cc|cc|cc|cc|cc|cc|cc}
            \toprule
            \multicolumn{1}{c|}{\multirow{3}{*}{\textbf{Model}}}
            & \multirow{3}{*}{\textbf{Size}} 
            & \multirow{3}{*}{\textbf{Frames}} 
            & \multicolumn{2}{c|}{\textbf{VS}}
            & \multicolumn{2}{c|}{\textbf{SC}}
            & \multicolumn{2}{c|}{\textbf{AU}}
            & \multicolumn{2}{c|}{\textbf{AG}}
            & \multicolumn{2}{c|}{\textbf{CO}}
            & \multicolumn{2}{c|}{\textbf{Overall}}
            & \multicolumn{2}{c|}{\textbf{BL-2}}
            & \multicolumn{2}{c|}{\textbf{MET}}
            & \multicolumn{2}{c}{\textbf{F1$_{\text{BERT}}$}} \\
            \cmidrule(lr){4-5} \cmidrule(lr){6-7} \cmidrule(lr){8-9}
            \cmidrule(lr){10-11} \cmidrule(lr){12-13} \cmidrule(lr){14-15}
            \cmidrule(lr){16-17} \cmidrule(lr){18-19} \cmidrule(lr){20-21}
            & &
            & w/o s. & w/ s. & w/o s. & w/ s. & w/o s. & w/ s. 
            & w/o s. & w/ s. & w/o s. & w/ s. & w/o s. & w/ s.
            & w/o s. & w/ s. & w/o s. & w/ s. & w/o s. & w/ s. \\
            \midrule
            \rowcolor{gray!20} 
            \multicolumn{21}{c}{\textit{Heuristics baselines}} \\
            \midrule
            Random Choice & - & - & \multicolumn{2}{c|}{39.3} & \multicolumn{2}{c|}{38.2} & \multicolumn{2}{c|}{35.5} & \multicolumn{2}{c|}{36.5} & \multicolumn{2}{c|}{38.8} & \multicolumn{2}{c|}{37.7} & - & - & - & - & - & - \\
            Frequent Guess & - & - & \multicolumn{2}{c|}{43.6} & \multicolumn{2}{c|}{46.0} & \multicolumn{2}{c|}{43.1} & \multicolumn{2}{c|}{41.1} & \multicolumn{2}{c|}{50.6} & \multicolumn{2}{c|}{44.4} & - & - & - & - & - & -\\
            Human & - & - & \multicolumn{2}{c|}{98.2} & \multicolumn{2}{c|}{94.4} & \multicolumn{2}{c|}{94.6} & \multicolumn{2}{c|}{97.2} & \multicolumn{2}{c|}{92.6} & \multicolumn{2}{c|}{95.8} & - & - & - & - & - & -\\
            
            \midrule
            \rowcolor{gray!20} 
            \multicolumn{21}{c}{\textit{Open-source Video MLLMs}} \\
            \midrule
            
            InternVL2\cite{internvl2} & 7B & 32 
            & \cellcolor{gray!30}50.2 & 52.0 & 51.3 & 56.8 & 52.6 & \cellcolor{gray!30}57.7 & \cellcolor{purple!30}73.6 & \cellcolor{orange!25}78.6 & \cellcolor{gray!30}55.1 & 57.7 & 55.2 & \cellcolor{gray!30}59.2 & \cellcolor{orange!25}10.11 & 8.43 & 23.68 & 21.22 & \cellcolor{orange!25}69.93 & 68.40\\
            
            LLaVA-OneVision\cite{llavaov} & 7B & 32 
            & 40.9 & 51.5 & 39.3 & 54.0 & 43.3 & 56.2 & 44.5 & 70.6 & 45.5 & \cellcolor{purple!30}59.6 & 42.0 & 56.9 & 6.94 & 7.30 & 19.67 & 20.20 & 67.03 & 67.82\\
            
            LLaVA-Video\cite{llavavideo} & 7B & 64 
            & 47.8 & \cellcolor{gray!30}54.9 & 48.7 & 56.8 & 46.7 & 57.4 & 50.2 & 71.2 & 49.4 & 51.9 & 48.3 & 58.3 & 6.50 & 7.27 & 17.35 & 19.06 & 66.59 & 67.95\\
            
            Qwen2-VL\cite{Qwen2VL} & 7B & 64 
            & \cellcolor{purple!30}52.9 & \cellcolor{orange!25}55.7 & \cellcolor{orange!25}55.0 & \cellcolor{orange!25}57.5 & \cellcolor{purple!30}56.4 & \cellcolor{orange!25}58.6 & \cellcolor{gray!30}72.2 & 75.3 & \cellcolor{purple!30}57.7 & \cellcolor{purple!30}59.6 & \cellcolor{purple!30}57.7 & \cellcolor{orange!25}60.3 & \cellcolor{purple!30}12.3 & \cellcolor{purple!30}11.41 & \cellcolor{orange!25}29.12 & \cellcolor{gray!30}27.97 & \cellcolor{purple!30}71.53 & \cellcolor{purple!30}70.71\\

            MiniCPM-V 2.6\cite{minicpm_yao2024} & 8B & 64 
            & 48.4 & 53.3 & \cellcolor{gray!30}53.1 & \cellcolor{gray!30}57.3 & \cellcolor{gray!30}54.7 & \cellcolor{gray!30}57.7 & \cellcolor{orange!25}73.2 & \cellcolor{gray!30}76.6 & \cellcolor{orange!25}56.4 & 55.1 & \cellcolor{gray!30}55.6 & 59.1 & \cellcolor{gray!30}9.39 & 8.57 & \cellcolor{purple!30}30.69 & \cellcolor{purple!30}30.06 & \cellcolor{gray!25}69.53 & 68.66\\
            
            Aria\cite{aria_li2024} & 8×3.5B & 128 
            & 48.2 & 52.4 & 50.1 & 56.6 & 45.3 & 51.8 & 64.2 & 75.6 & 46.8 & 50.0 & 50.5 & 56.8 & 9.25 & \cellcolor{orange!25}10.00 & 23.25 & 26.08 & 69.10 & \cellcolor{orange!30}70.35\\
            
            VITA\cite{vita_fu2024} & 8×7B & 32 
            & 41.6 & 46.9 & 44.1 & 46.0 & 47.2 & 53.3 & 59.9 & 68.9 & 42.9 & 48.1 & 46.5 & 51.8 & 8.65 & 9.32 & 25.06 & 27.08 & 68.81 & 68.93\\
            
            \midrule
            \rowcolor{gray!20} 
            \multicolumn{21}{c}{\textit{Open-source Video-Audio MLLMs}} \\
            \midrule
            
            % need modification
            % \multirow{2}{*}{\makecell{VideoLLaMA2.1-AV}}

            VideoLLaMA2.1-AV\cite{videollama2_damonlpsg2024} & \multirow{2}{*}{7B} & \multirow{2}{*}{32} 
            & 41.2 & 47.4 & 42.3 & 52.2 & 43.3 & 53.3 & 36.8 & 66.6 & 35.9 & 49.4 & 40.8 & 53.1 & 6.68 & 7.61 & 21.67 & 23.34 & 66.02 & 67.53\\
            \multicolumn{1}{r|}{+audio} & & 
            & 39.1 & 45.3 & 40.9 & 52.2 & 42.6 & 51.3 & 34.8 & 66.2 & 32.1 & 46.2 & 39.0 & 51.7 & 6.17 & 7.24 & 20.25 & 22.71 & 64.85 & 67.09\\
            
            \midrule
            \rowcolor{gray!20} 
            \multicolumn{21}{c}{\textit{Closed-source MLLMs}} \\
            \midrule

            GPT-4o-mini\cite{gpt4o_mini_openai} & N/A & 50 & 40.1 & 46.7 & 34.4 & 42.7 & 40.4 & 47.9 & 54.5 & 70.9 & 41.0 & 44.9 & 41.3 & 49.8 & 9.34 & \cellcolor{gray!30}9.99 & \cellcolor{gray!30}26.79 & \cellcolor{orange!25}29.30 & 68.35 & 68.76\\
            
            GPT-4o\cite{gpt4o_openai} & N/A & 50 & \cellcolor{orange!25}52.4 & \cellcolor{purple!30}55.8 & \cellcolor{purple!30}56.4 & \cellcolor{purple!30}62.8 & \cellcolor{orange!25}56.0 & \cellcolor{purple!30}60.6 & \cellcolor{gray!30}72.2 & \cellcolor{purple!30}79.9 & 47.4 & \cellcolor{purple!30}59.6 & \cellcolor{orange!25}56.9 & \cellcolor{purple!30}62.8 & 7.45 & 9.61 & 22.02 & 25.10 &  67.74 & \cellcolor{gray!30}68.94\\
            
            \bottomrule
        \end{tabular}
    }
    \vspace{-5pt}
    
    \caption{
        \textbf{Performance of MLLMs on {\BenchName}.}
        Size means the LLM size.
        Judgement and multichoice metrics Accuracy and open-ended metrics BLEU-2(\textbf{BL-2}), METEOR (\textbf{MET}), and BERTScore F1 (\textbf{F1$_{\text{BERT}}$}) are reported in percentage (\%), evaluated under two settings: ``without subtitles'' (\textbf{w/o s.}) and ``with subtitles'' (\textbf{w/ s.}).
        % Overall evaluation scores(\textbf{Overall}) are calculated by averaging the scores of judgment and multichoice metrics.
        ``-'' indicates that results are not feasible with open-ended metrics in heuristic baselines.
        The best, second-best, and third-best results are marked \colorbox{purple!30}{purple}, \colorbox{orange!25}{orange}, and \colorbox{gray!30}{gray}, respectively.
    }
    \vspace{-0.5cm} % 调整间距，负值表示减少间距
    \label{tab:main_results_sub}
\end{table*}

% \cellcolor{purple!30}    % top 1 - 浅紫色
% \cellcolor{orange!25}    % top 2 - 浅橙色
% \cellcolor{gray!30}      % top 3 - 浅灰色

\begin{table*}[!htp]
    \centering
    \small
    \vspace{10pt} % 调整间距，负值表示减少间距
    \renewcommand\tabcolsep{8pt} % column space
    \renewcommand\arraystretch{0.80} % row space
    \resizebox{1.0\linewidth}{!}{
        \begin{tabular}{l|c|c|c|c|c|c|c|c|c|c}
            \toprule
            % header of table
            \rowcolor{gray!25} 
            \multicolumn{1}{c|}{\textbf{Model}}
            & \multicolumn{1}{c|}{\textbf{Episodes}}
            & \multicolumn{1}{c|}{\textbf{VS}}
            & \multicolumn{1}{c|}{\textbf{SC}}
            & \multicolumn{1}{c|}{\textbf{AU}}
            & \multicolumn{1}{c|}{\textbf{AG}}
            & \multicolumn{1}{c|}{\textbf{CO}}
            & \textbf{Overall}
            & \textbf{BL-2}
            & \textbf{MET}
            & \textbf{F1$_{\text{BERT}}$} \\

            \midrule

            \multirow{5}{*}{InternVL2}
            & - & 52.0 & 56.8 & 57.7 & 78.6 & 57.7 & \textbf{59.2} & 8.43 & 21.22 & 68.40 \\ 
            & $\text{Prev}_2$ & 50.7 & 55.2 & 58.4 & 73.6 & 58.3 & \textbf{57.8} & 9.66 & 23.16 & 69.36 \\ 
            & $\text{Prev}_1$ & 50.5 & 57.7 & 59.1 & 73.2 & 57.1 & \textbf{58.4} & 9.37 & 22.1 & 68.81 \\ 
            & $\text{Next}_1$ & 50.9 & 55.4 & 57.4 & 76.9 & 59.6 & \textbf{58.4} & 8.81 & 21.57 & 68.67 \\ 
            & $\text{Next}_2$ & 49.1 & 56.4 & 58.4 & 73.9 & 57.1 & \textbf{57.6} & 8.33 & 20.89 & 68.48 \\ 

            \midrule
            
            \multirow{5}{*}{Qwen2-VL}
            & - & 55.7 & 57.5 & 58.6 & 75.3 & 59.6 & \textbf{60.3} & 11.41 & 27.97 & 70.71 \\
            & $\text{Prev}_2$ & 51.6 & 58.2 & 58.6 & 71.2 & 60.9 & \textbf{58.7} & 10.58 & 27.02 & 70.01 \\
            & $\text{Prev}_2$ & 54.9 & 58.0 & 58.4 & 71.2 & 59.0 & \textbf{59.4} & 10.91 & 27.35 & 70.36 \\
            & $\text{Next}_1$ & 54.6 & 58.9 & 56.9 & 72.2 & 59.0 & \textbf{59.3} & 10.83 & 27.13 & 70.39 \\
            & $\text{Next}_2$ & 52.9 & 58.0 & 56.0 & 72.2 & 59.6 & \textbf{58.5} & 10.78 & 26.91 & 70.28 \\

            \midrule
            
            \multirow{5}{*}{MiniCPM-V 2.6}
            & - & 53.3 & 57.3 & 57.7 & 76.6 & 55.1 & \textbf{59.1} & 8.57 & 30.06 & 68.66 \\
            & $\text{Prev}_2$ & 48.7 & 52.2 & 56.9 & 69.2 & 50.0 & \textbf{54.8} & 7.88 & 29.26 & 68.33 \\
            & $\text{Prev}_1$ & 48.0 & 52.0 & 55.5 & 71.2 & 53.2 & \textbf{54.8} & 7.77 & 29.52 & 68.42 \\
            & $\text{Next}_1$ & 48.2 & 52.2 & 50.4 & 70.2 & 48.1 & \textbf{53.2} & 7.97 & 30.01 & 68.32 \\
            & $\text{Next}_2$ & 46.4 & 52.2 & 54.0 & 61.2 & 48.7 & \textbf{52.0} & 7.63 & 29.47 & 68.17 \\
            \bottomrule
            
        \end{tabular}
    }
    \vspace{-5pt} % 调整间距，负值表示减少间距
    % 这里给出了总表中效果比较好的几个模型的结果，包含InternVL2, Qwen2-VL, MiniCPM-V 2.6, GPT-4o
    \caption{
        \textbf{Performance of Top-Performing MLLMs on Multi-Episode Series Tasks.}
    }
    \vspace{-0.5cm} % 调整间距，负值表示减少间距

    \label{tab:series_results_all}
\end{table*}

\subsection{Performance on Multi-Episode Tasks}
Tab.~\ref{tab:series_results_all} presents the performance of various models on multi-episode tasks. We observe that model performance with multi-episode inputs is consistently lower compared to single-episode inputs, highlighting the difficulty in comprehending narrative structures and character dynamics within series. Addressing these limitations requires advancements in modeling long-term dependencies and context-aware reasoning, which remain critical areas for future research.

\subsection{Fine-Grained Task Results}
Tab. \ref{tab:leaderboard} presents more results of different fine-grained tasks in {\BenchName}. For the tasks that are strongly related to the plot (a) (b) (c) (d) (e) (f) and those strongly related to characters (i) (u) (s) (t), the top three models in most cases utilize the PC-DCoT framework. This demonstrates the framework's effectiveness in handling tasks closely related to narrative and plot elements, showcasing its ability to capture intricate relationships within the storyline, manage complex contextual dependencies, and enhance comprehension of plot progression. 

\subsection{Qualitative Results}
\begin{figure*}[thp]
    \centering
    % \vspace{-0.3cm}
    \includegraphics[width=1.0\textwidth]{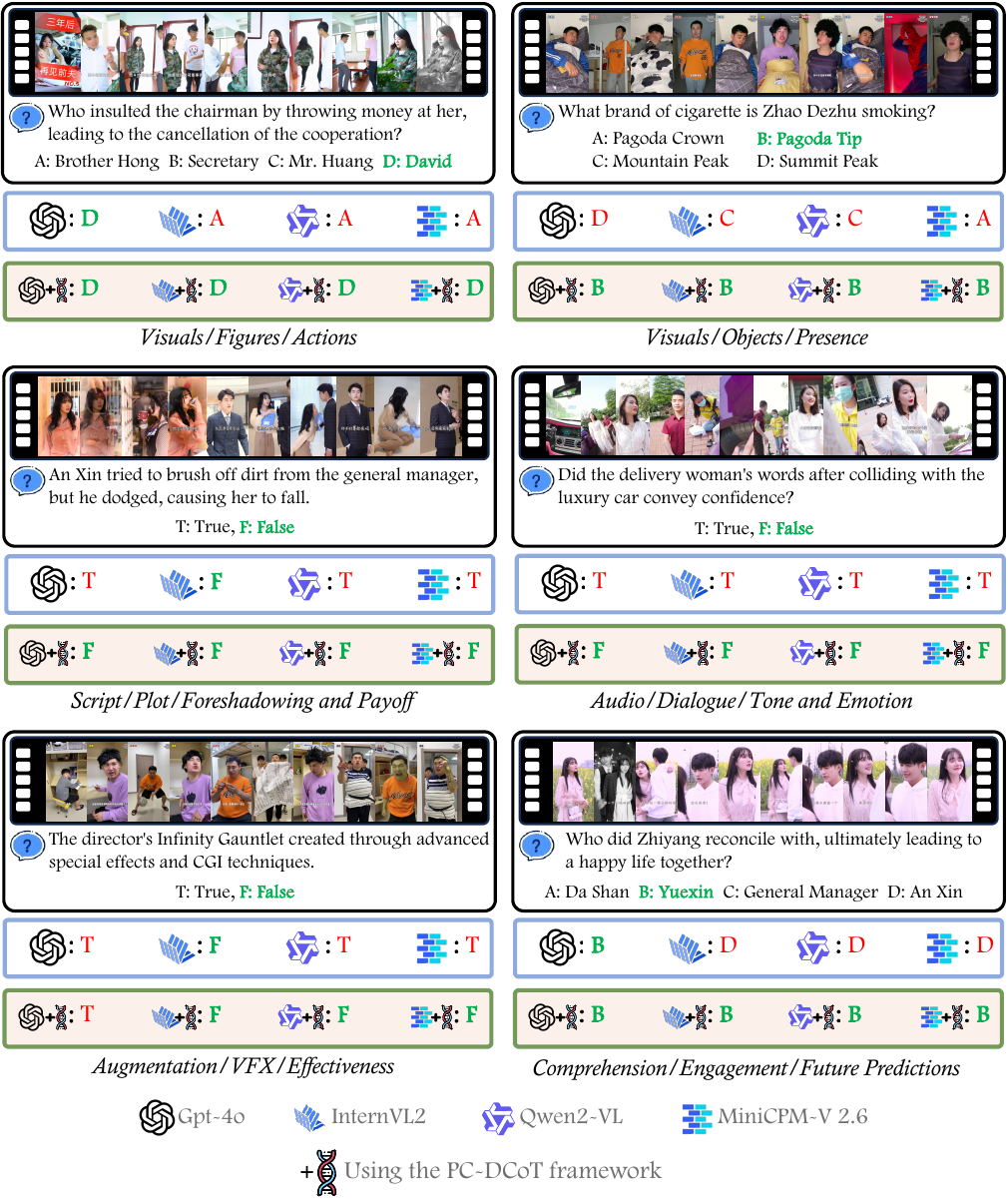}
    \vspace{-10px}
    \caption{
    \textbf{More qualitative cases in SeriesBench.} 
      The performance of MLLM can be improved by using the PC-DCoT framework. We use \textcolor[RGB]{0,100,0}{Green} to indicate correct and \textcolor{red}{Red} to indicate incorrect.
    }
    \label{fig:qualitative}
    \vspace{-0.3cm}
\end{figure*}

\begin{figure*}[thp]
    \centering
    % \vspace{-0.3cm}
    \includegraphics[width=1.0\textwidth]{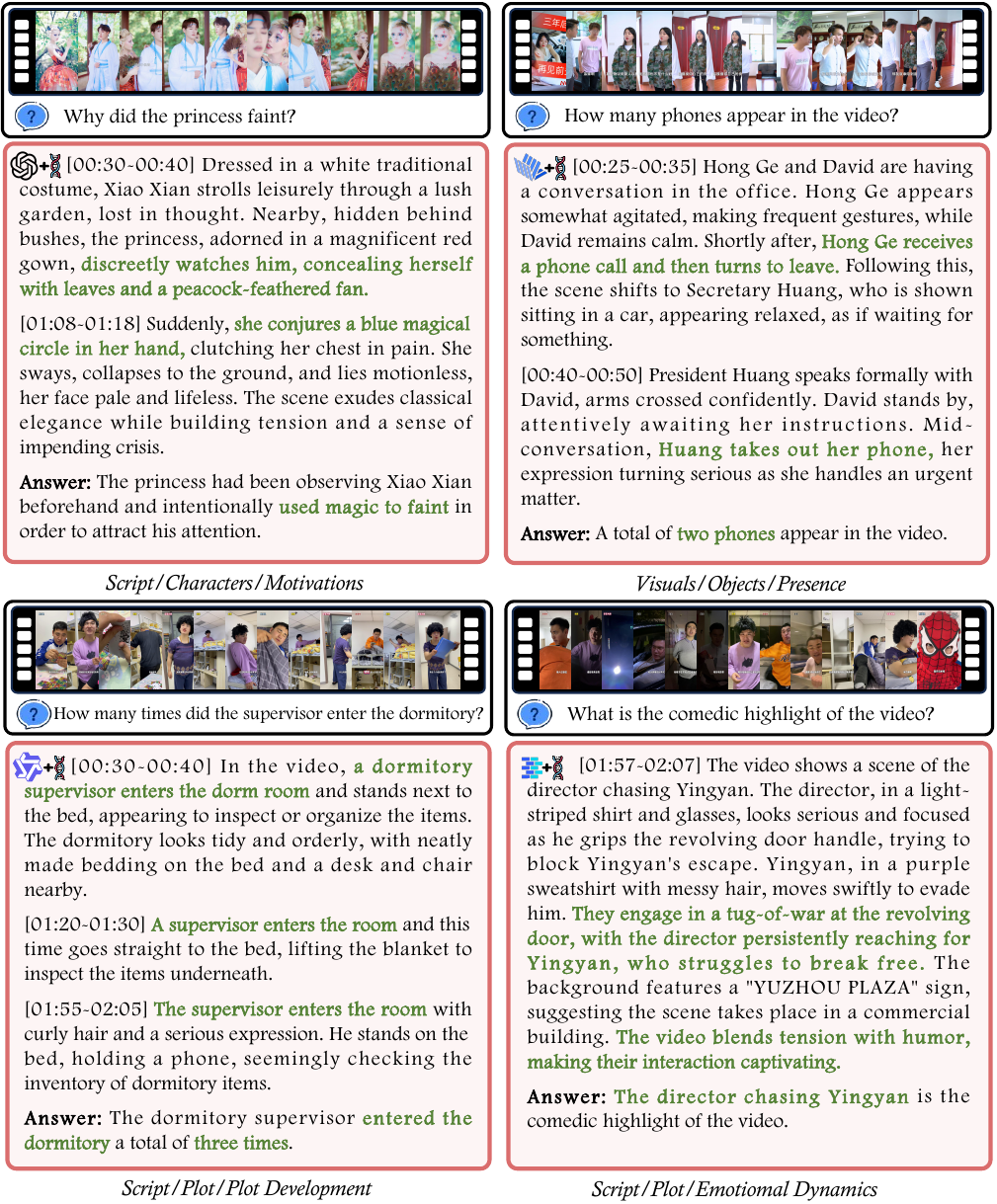}
    \vspace{-10px}
    \caption{
    \textbf{Additional examples of open-ended responses using PC-DCoT.} 
       MLLMs are capable of conducting more detailed analyses of events, relationships, and individuals, demonstrating a deeper understanding of narrative-driven series.
    }
    \vspace{-5px}
    \label{fig:qualitative_openend}
    \vspace{-0.3cm}
\end{figure*}

More additional qualitative results can be found in Figs. \ref{fig:qualitative} and \ref{fig:qualitative_openend}. Figs. \ref{fig:qualitative} showcases model reasoning performance on multiple-choice and true/false tasks across different task dimensions. These examples highlight the necessity for narrative comprehension when tackling questions within our SeriesBench. It is evident that even state-of-the-art Multimodal Large Language Models (MLLMs) struggle to provide correct answers for certain questions, underscoring the complexity of the tasks. However, by incorporating the PC-DCoT framework, model performance shows notable improvement, demonstrating enhanced comprehension and reasoning capabilities when addressing narrative-driven tasks.

Fig. \ref{fig:qualitative_openend} illustrates the intermediate reasoning process and final answers of MLLMs when addressing open-ended questions using the PC-DCoT framework. The results reveal that integrating the PC-DCoT framework enables MLLMs to perform deeper analyses of events and individuals, enhancing their ability to draw nuanced conclusions. This improvement allows the model to tackle more complex problems effectively, demonstrating the framework's capability to systematically guide reasoning processes toward more accurate and comprehensive outcomes.

\begin{table*}[tp]
\vspace{-4px}
\centering
\begin{minipage}[t]{0.22\textwidth} 
    \centering
    \setlength{\tabcolsep}{3.0pt}
    \resizebox{1\linewidth}{!}{%
        \begin{tabular}{c|l|c}
            \textbf{Rank} & \multicolumn{1}{c|}{\textbf{Model}} & \textbf{Score} \\
            \Xhline{1.0pt}
                    \rowcolor{red!20}\faTrophy & \textbf{GPT-4o+PC-DCoT} & \textbf{79.1} \\
            \rowcolor{red!12}\textbf{2} & \textbf{InternVL2+PC-DCoT} & \textbf{74.4} \\
            \rowcolor{red!6}\textbf{3} & \textbf{Qwen2-VL+PC-DCoT} & \textbf{72.1} \\
            4 & InternVL2 & 65.3 \\
            5 & MiniCPM-V 2.6+PC-DCoT & 65.1 \\
            6 & MiniCPM-V 2.6 & 63.3 \\
            7 & GPT-4o & 59.2 \\
            8 & LLaVA-OneVision & 57.1 \\
            9 & Qwen2-VL & 55.1 \\
            10 & Aria & 53.1 \\
            11 & LLaVA-Video & 42.9 \\
            12 & VideoLLaMA2.1-AV & 42.9 \\
        \end{tabular}
    }
    \subcaption{\textit{Foreshadowing and Payoff}}
    \vspace{-2px}
\end{minipage}
\hfill
\begin{minipage}[t]{0.22\textwidth} 
    
    \centering
    \setlength{\tabcolsep}{3.0pt}
    \resizebox{1\linewidth}{!}{%
        \begin{tabular}{c|l|c}
            \textbf{Rank} & \multicolumn{1}{c|}{\textbf{Model}} & \textbf{Score} \\
            \Xhline{1.0pt}
                    \rowcolor{red!20}\faTrophy & \textbf{InternVL2+PC-DCoT} & \textbf{80.0} \\
            \rowcolor{red!12}\textbf{2} & \textbf{Qwen2-VL+PC-DCoT} & \textbf{80.0} \\
            \rowcolor{red!6}\textbf{3} & \textbf{GPT-4o+PC-DCoT} & \textbf{75.0} \\
            4 & MiniCPM-V 2.6+PC-DCoT & 75.0 \\
            5 & InternVL2 & 71.4 \\
            6 & GPT-4o & 71.4 \\
            7 & Qwen2-VL & 66.7 \\
            8 & MiniCPM-V 2.6 & 61.9 \\
            9 & LLaVA-OneVision & 61.9 \\
            10 & VideoLLaMA2.1-AV & 57.1 \\
            11 & Aria & 57.1 \\
            12 & LLaVA-Video & 42.9 \\
       
        \end{tabular}
    }
    \subcaption{ \textit{Suspense and Continuity}}
    \vspace{-2px}
\end{minipage}
\hfill
\begin{minipage}[t]{0.22\textwidth} 
    
    \centering
    \setlength{\tabcolsep}{3.0pt}
    \resizebox{1\linewidth}{!}{%
        \begin{tabular}{c|l|c}
            \textbf{Rank} & \multicolumn{1}{c|}{\textbf{Model}} & \textbf{Score} \\
            \Xhline{1.0pt}
                    \rowcolor{red!20}\faTrophy & \textbf{Qwen2-VL+PC-DCoT} & \textbf{75.0} \\
            \rowcolor{red!12}\textbf{2} & \textbf{GPT-4o} & \textbf{67.9} \\
            \rowcolor{red!6}\textbf{3} & \textbf{InternVL2+PC-DCoT} & \textbf{64.3} \\
            4 & GPT-4o+PC-DCoT & 64.3 \\
            5 & MiniCPM-V 2.6+PC-DCoT & 64.3 \\
            6 & LLaVA-Video & 50.0 \\
            7 & Aria & 50.0 \\
            8 & VideoLLaMA2.1-AV & 46.4 \\
            9 & Qwen2-VL & 46.4 \\
            10 & MiniCPM-V 2.6 & 46.4 \\
            11 & InternVL2 & 46.4 \\
            12 & LLaVA-OneVision & 39.3 \\
        
        \end{tabular}
    }
    \subcaption{ \textit{Emotional Dynamics}}
    \vspace{-2px}
\end{minipage}
\hfill
\begin{minipage}[t]{0.22\textwidth} 
    
    \centering
    \setlength{\tabcolsep}{3.0pt}
    \resizebox{1\linewidth}{!}{%
        \begin{tabular}{c|l|c}
            \textbf{Rank} & \multicolumn{1}{c|}{\textbf{Model}} & \textbf{Score} \\
            \Xhline{1.0pt}
                    \rowcolor{red!20}\faTrophy & \textbf{InternVL2+PC-DCoT} & \textbf{76.9} \\
            \rowcolor{red!12}\textbf{2} & \textbf{MiniCPM-V 2.6+PC-DCoT} & \textbf{76.9} \\
            \rowcolor{red!6}\textbf{3} & \textbf{Qwen2-VL+PC-DCoT} & \textbf{73.1} \\
            
            4 & GPT-4o & 67.9 \\
            5 & GPT-4o+PC-DCoT & 65.4 \\
            6 & LLaVA-OneVision & 64.3 \\
            7 & VideoLLaMA2.1-AV & 60.7 \\
            8 & MiniCPM-V 2.6 & 60.7 \\
            9 & Aria & 60.7 \\
            10 & InternVL2 & 57.1 \\
            11 & LLaVA-Video & 53.6 \\
            12 & Qwen2-VL & 53.6 \\
        \end{tabular}
    }
    \subcaption{ \textit{Twists and Conflicts}}
    \vspace{-2px}
\end{minipage}

\vspace{0.1cm}

\begin{minipage}[t]{0.22\textwidth} 
    
    \centering
    \setlength{\tabcolsep}{3.0pt}
    \resizebox{1\linewidth}{!}{%
        \begin{tabular}{c|l|c}
            \textbf{Rank} & \multicolumn{1}{c|}{\textbf{Model}} & \textbf{Score} \\
            \Xhline{1.0pt}
                    \rowcolor{red!20}\faTrophy & \textbf{Qwen2-VL+PC-DCoT} & \textbf{75.0} \\
            \rowcolor{red!12}\textbf{2} & \textbf{GPT-4o+PC-DCoT} & \textbf{71.9} \\
            \rowcolor{red!6}\textbf{3} & \textbf{LLaVA-Video} & \textbf{71.4} \\
            4 & VideoLLaMA2.1-AV & 71.4 \\
            5 & GPT-4o & 71.4 \\
            6 & InternVL2+PC-DCoT & 68.8 \\
            7 & Qwen2-VL & 68.6 \\
            8 & Aria & 68.6 \\
            9 & MiniCPM-V 2.6+PC-DCoT & 59.4 \\
            10 & MiniCPM-V 2.6 & 54.3 \\
            11 & LLaVA-OneVision & 54.3 \\
            12 & InternVL2 & 51.4 \\
        
        \end{tabular}
    }
    \subcaption{ \textit{Climaxes and Build-ups}}
    \vspace{-2px}
\end{minipage}
\hfill
\begin{minipage}[t]{0.22\textwidth} 
    
    \centering
    \setlength{\tabcolsep}{3.0pt}
    \resizebox{1\linewidth}{!}{%
        \begin{tabular}{c|l|c}
            \textbf{Rank} & \multicolumn{1}{c|}{\textbf{Model}} & \textbf{Score} \\
            \Xhline{1.0pt}
                    \rowcolor{red!20}\faTrophy & \textbf{GPT-4o+PC-DCoT} & \textbf{78.3} \\
            \rowcolor{red!12}\textbf{2} & \textbf{MiniCPM-V 2.6+PC-DCoT} & \textbf{69.6} \\
            \rowcolor{red!6}\textbf{3} & \textbf{InternVL2+PC-DCoT} & \textbf{67.8} \\
            4 & Qwen2-VL+PC-DCoT & 67.8 \\
            5 & LLaVA-Video & 60.3 \\
            6 & GPT-4o & 59.5 \\
            7 & MiniCPM-V 2.6 & 53.7 \\
            8 & Aria & 50.4 \\
            9 & LLaVA-OneVision & 50.4 \\
            10 & InternVL2 & 49.6 \\
            11 & VideoLLaMA2.1-AV & 47.1 \\
            12 & Qwen2-VL & 46.3 \\
        
        \end{tabular}
    }
    \subcaption{ \textit{Plot Development}}
    \vspace{-2px}
\end{minipage}
\hfill
\begin{minipage}[t]{0.22\textwidth} 
    
    \centering
    \setlength{\tabcolsep}{3.0pt}
    \resizebox{1\linewidth}{!}{%
        \begin{tabular}{c|l|c}
            \textbf{Rank} & \multicolumn{1}{c|}{\textbf{Model}} & \textbf{Score} \\
            \Xhline{1.0pt}
                    \rowcolor{red!20}\faTrophy & \textbf{Qwen2-VL+PC-DCoT} & \textbf{76.5} \\
            \rowcolor{red!12}\textbf{2} & \textbf{Qwen2-VL} & \textbf{75.0} \\
            \rowcolor{red!6}\textbf{3} & \textbf{LLaVA-Video} & \textbf{70.0} \\
       
            4 & MiniCPM-V 2.6 & 65.0 \\
            5 & Aria & 65.0 \\
            6 & GPT-4o+PC-DCoT & 64.7 \\
            7 & InternVL2+PC-DCoT & 58.8 \\
            8 & MiniCPM-V 2.6+PC-DCoT & 58.8 \\
            9 & VideoLLaMA2.1-AV & 55.0 \\
            10 & LLaVA-OneVision & 55.0 \\
            11 & GPT-4o & 50.0 \\
            12 & InternVL2 & 45.0 \\
        \end{tabular}
    }
    \subcaption{ \textit{World-Building}}
    \vspace{-2px}
\end{minipage}
\hfill
\begin{minipage}[t]{0.22\textwidth} 
    
    \centering
    \setlength{\tabcolsep}{3.0pt}
    \resizebox{1\linewidth}{!}{%
        \begin{tabular}{c|l|c}
            \textbf{Rank} & \multicolumn{1}{c|}{\textbf{Model}} & \textbf{Score} \\
            \Xhline{1.0pt}
                    \rowcolor{red!20}\faTrophy & \textbf{Qwen2-VL} & \textbf{78.9} \\
            \rowcolor{red!12}\textbf{2} & \textbf{GPT-4o+PC-DCoT} & \textbf{75.0} \\
            \rowcolor{red!6}\textbf{3} & \textbf{LLaVA-Video} & \textbf{73.7} \\
            4 & Aria & 73.7 \\
            5 & Qwen2-VL+PC-DCoT & 68.8 \\
            6 & LLaVA-OneVision & 68.4 \\
            7 & VideoLLaMA2.1-AV & 63.2 \\
            8 & InternVL2 & 57.9 \\
         
            9 & GPT-4o & 57.9 \\
            10 & InternVL2+PC-DCoT & 56.2 \\
            11 & MiniCPM-V 2.6+PC-DCoT & 56.2 \\
            12 & MiniCPM-V 2.6 & 52.6 \\
        \end{tabular}
    }
    \subcaption{ \textit{Time and Location}}
    \vspace{-2px}
\end{minipage}
\hfill

\vspace{0.1cm}

\begin{minipage}[t]{0.22\textwidth} 
    
    \centering
    \setlength{\tabcolsep}{3.0pt}
    \resizebox{1\linewidth}{!}{%
        \begin{tabular}{c|l|c}
            \textbf{Rank} & \multicolumn{1}{c|}{\textbf{Model}} & \textbf{Score} \\
            \Xhline{1.0pt}
                    \rowcolor{red!20}\faTrophy & \textbf{GPT-4o+PC-DCoT} & \textbf{80.0} \\
            \rowcolor{red!12}\textbf{2} & \textbf{InternVL2+PC-DCoT} & \textbf{78.6} \\
            \rowcolor{red!6}\textbf{3} & \textbf{Qwen2-VL+PC-DCoT} & \textbf{75.7} \\
            4 & InternVL2 & 71.8 \\
            5 & MiniCPM-V 2.6+PC-DCoT & 70.0 \\
            6 & Qwen2-VL & 69.0 \\
            7 & GPT-4o & 69.0 \\
            8 & Aria & 67.6 \\
            9 & LLaVA-OneVision & 66.2 \\
            10 & LLaVA-Video & 64.8 \\
            11 & MiniCPM-V 2.6 & 60.6 \\
            12 & VideoLLaMA2.1-AV & 59.2 \\
        
        \end{tabular}
    }
    \subcaption{ \textit{Character Motivations}}
    \vspace{-2px}
\end{minipage}
\hfill
\begin{minipage}[t]{0.22\textwidth} 
    
    \centering
    \setlength{\tabcolsep}{3.0pt}
    \resizebox{1\linewidth}{!}{%
        \begin{tabular}{c|l|c}
            \textbf{Rank} & \multicolumn{1}{c|}{\textbf{Model}} & \textbf{Score} \\
            \Xhline{1.0pt}
                    \rowcolor{red!20}\faTrophy & \textbf{GPT-4o+PC-DCoT} & \textbf{71.4} \\
            \rowcolor{red!12}\textbf{2} & \textbf{Qwen2-VL+PC-DCoT} & \textbf{66.7} \\
            \rowcolor{red!6}\textbf{3} & \textbf{InternVL2+PC-DCoT} & \textbf{60.3} \\
            4 & MiniCPM-V 2.6+PC-DCoT & 57.1 \\
            5 & InternVL2 & 52.9 \\
            6 & Qwen2-VL & 50.0 \\
            7 & LLaVA-Video & 48.5 \\
            8 & MiniCPM-V 2.6 & 48.5 \\
            9 & LLaVA-OneVision & 48.5 \\
            10 & GPT-4o & 47.1 \\
            11 & Aria & 45.6 \\
            12 & VideoLLaMA2.1-AV & 39.7 \\
            
        \end{tabular}
    }
    \subcaption{\textit{Pronoun References}}
    \vspace{-2px}
\end{minipage}
\hfill
\begin{minipage}[t]{0.22\textwidth} 
    
    \centering
    \setlength{\tabcolsep}{3.0pt}
    \resizebox{1\linewidth}{!}{%
        \begin{tabular}{c|l|c}
            \textbf{Rank} & \multicolumn{1}{c|}{\textbf{Model}} & \textbf{Score} \\
            \Xhline{1.0pt}
                    \rowcolor{red!20}\faTrophy & \textbf{MiniCPM-V 2.6+PC-DCoT} & \textbf{85.8} \\
            \rowcolor{red!12}\textbf{2} & \textbf{Qwen2-VL+PC-DCoT} & \textbf{85.2} \\
            \rowcolor{red!6}\textbf{3} & \textbf{GPT-4o} & \textbf{84.6} \\
            4 & GPT-4o+PC-DCoT & 84.2 \\
            5 & InternVL2+PC-DCoT & 83.6 \\
            6 & MiniCPM-V 2.6 & 83.1 \\
            7 & InternVL2 & 83.1 \\
            8 & Qwen2-VL & 82.1 \\
            9 & Aria & 81.5 \\
           
            10 & LLaVA-Video & 79.0 \\
            11 & LLaVA-OneVision & 78.5 \\
            12 & VideoLLaMA2.1-AV & 72.8 \\
        \end{tabular}
    }
    \subcaption{ \textit{Subtitles Recognition}}
    \vspace{-2px}
\end{minipage}
\hfill
\begin{minipage}[t]{0.22\textwidth} 
    
    \centering
    \setlength{\tabcolsep}{3.0pt}
    \resizebox{1\linewidth}{!}{%
        \begin{tabular}{c|l|c}
            \textbf{Rank} & \multicolumn{1}{c|}{\textbf{Model}} & \textbf{Score} \\
            \Xhline{1.0pt}
                    \rowcolor{red!20}\faTrophy & \textbf{Qwen2-VL+PC-DCoT} & \textbf{94.7} \\
            \rowcolor{red!12}\textbf{2} & \textbf{GPT-4o+PC-DCoT} & \textbf{94.7} \\
            \rowcolor{red!6}\textbf{3} & \textbf{MiniCPM-V 2.6+PC-DCoT} & \textbf{94.7} \\
            4 & InternVL2 & 89.5 \\
            5 & InternVL2+PC-DCoT & 89.5 \\
            6 & GPT-4o & 84.2 \\
            7 & Qwen2-VL & 78.9 \\
            8 & MiniCPM-V 2.6 & 73.7 \\
            9 & VideoLLaMA2.1-AV & 57.9 \\
            10 & Aria & 52.6 \\
            11 & LLaVA-OneVision & 52.6 \\
            12 & LLaVA-Video & 47.4 \\
  
        \end{tabular}
    }
    \subcaption{ \textit{Labels Purpose}}
    \vspace{-2px}
\end{minipage}
\hfill

\vspace{0.1cm}

\begin{minipage}[t]{0.22\textwidth} 
    
    \centering
    \setlength{\tabcolsep}{3.0pt}
    \resizebox{1\linewidth}{!}{%
        \begin{tabular}{c|l|c}
            \textbf{Rank} & \multicolumn{1}{c|}{\textbf{Model}} & \textbf{Score} \\
            \Xhline{1.0pt}
                    \rowcolor{red!20}\faTrophy & \textbf{GPT-4o+PC-DCoT} & \textbf{76.4} \\
            \rowcolor{red!12}\textbf{2} & \textbf{MiniCPM-V 2.6+PC-DCoT} & \textbf{76.4} \\
            \rowcolor{red!6}\textbf{3} & \textbf{InternVL2+PC-DCoT} & \textbf{75.0} \\
            4 & Qwen2-VL+PC-DCoT & 70.8 \\
            5 & GPT-4o & 70.1 \\
            6 & Aria & 66.2 \\
            7 & Qwen2-VL & 63.6 \\
            8 & MiniCPM-V 2.6 & 63.6 \\
            9 & InternVL2 & 62.3 \\
            10 & LLaVA-Video & 58.4 \\
            11 & LLaVA-OneVision & 58.4 \\
            12 & VideoLLaMA2.1-AV & 54.5 \\
   
        \end{tabular}
    }
    \subcaption{ \textit{Effectiveness}}
    \vspace{-2px}
\end{minipage}
\hfill
\begin{minipage}[t]{0.22\textwidth} 
    
    \centering
    \setlength{\tabcolsep}{3.0pt}
    \resizebox{1\linewidth}{!}{%
        \begin{tabular}{c|l|c}
            \textbf{Rank} & \multicolumn{1}{c|}{\textbf{Model}} & \textbf{Score} \\
            \Xhline{1.0pt}
                    \rowcolor{red!20}\faTrophy & \textbf{GPT-4o+PC-DCoT} & \textbf{82.7} \\
            \rowcolor{red!12}\textbf{2} & \textbf{InternVL2+PC-DCoT} & \textbf{78.0} \\
            \rowcolor{red!6}\textbf{3} & \textbf{MiniCPM-V 2.6+PC-DCoT} & \textbf{78.0} \\
            4 & Qwen2-VL+PC-DCoT & 77.2 \\
            5 & LLaVA-Video & 64.0 \\
            6 & MiniCPM-V 2.6 & 63.2 \\
            7 & Aria & 61.8 \\
            8 & InternVL2 & 57.4 \\
            9 & LLaVA-OneVision & 57.4 \\
            10 & VideoLLaMA2.1-AV & 56.6 \\
            11 & Qwen2-VL & 56.6 \\
            12 & GPT-4o & 55.9 \\
        \end{tabular}
    }
    \subcaption{\textit{Object Presence}}
    \vspace{-2px}
\end{minipage}
\hfill
\begin{minipage}[t]{0.22\textwidth} 
    
    \centering
    \setlength{\tabcolsep}{3.0pt}
    \resizebox{1\linewidth}{!}{%
        \begin{tabular}{c|l|c}
            \textbf{Rank} & \multicolumn{1}{c|}{\textbf{Model}} & \textbf{Score} \\
            \Xhline{1.0pt}
                    \rowcolor{red!20}\faTrophy & \textbf{InternVL2+PC-DCoT} & \textbf{81.8} \\
            \rowcolor{red!12}\textbf{2} & \textbf{GPT-4o+PC-DCoT} & \textbf{81.8} \\
            \rowcolor{red!6}\textbf{3} & \textbf{MiniCPM-V 2.6+PC-DCoT} & \textbf{79.2} \\
            4 & Qwen2-VL+PC-DCoT & 77.9 \\
            5 & MiniCPM-V 2.6 & 62.0 \\
            6 & InternVL2 & 60.8 \\
            7 & LLaVA-OneVision & 60.8 \\
            8 & GPT-4o & 60.8 \\
            9 & VideoLLaMA2.1-AV & 59.5 \\
            10 & LLaVA-Video & 54.4 \\
            11 & Qwen2-VL & 54.4 \\
            12 & Aria & 54.4 \\
        \end{tabular}
    }
    \subcaption{\textit{Dialogue Attribution}}
    \vspace{-2px}
\end{minipage}
\hfill
\begin{minipage}[t]{0.22\textwidth} 
    
    \centering
    \setlength{\tabcolsep}{3.0pt}
    \resizebox{1\linewidth}{!}{%
        \begin{tabular}{c|l|c}
            \textbf{Rank} & \multicolumn{1}{c|}{\textbf{Model}} & \textbf{Score} \\
            \Xhline{1.0pt}
                    \rowcolor{red!20}\faTrophy & \textbf{LLaVA-Video} & \textbf{59.1} \\
            \rowcolor{red!12}\textbf{2} & \textbf{VideoLLaMA2.1-AV} & \textbf{50.0} \\
            \rowcolor{red!6}\textbf{3} & \textbf{GPT-4o} & \textbf{50.0} \\
            4 & InternVL2+PC-DCoT & 47.6 \\
            5 & GPT-4o+PC-DCoT & 47.6 \\
            6 & Aria & 45.5 \\
            7 & LLaVA-OneVision & 45.5 \\
            8 & MiniCPM-V 2.6+PC-DCoT & 42.9 \\
            9 & InternVL2 & 36.4 \\
          
            10 & Qwen2-VL+PC-DCoT & 33.3 \\
            11 & MiniCPM-V 2.6 & 31.8 \\
            12 & Qwen2-VL & 27.3 \\
        \end{tabular}
    }
    \subcaption{ \textit{Character Resonance}}
    \vspace{-2px}
\end{minipage}

\vspace{0.1cm}

\begin{minipage}[t]{0.22\textwidth} 
    
    \centering
    \setlength{\tabcolsep}{3.0pt}
    \resizebox{1\linewidth}{!}{%
        \begin{tabular}{c|l|c}
            \textbf{Rank} & \multicolumn{1}{c|}{\textbf{Model}} & \textbf{Score} \\
            \Xhline{1.0pt}
                    \rowcolor{red!20}\faTrophy & \textbf{Qwen2-VL} & \textbf{70.6} \\
            \rowcolor{red!12}\textbf{2} & \textbf{InternVL2+PC-DCoT} & \textbf{68.8} \\
            \rowcolor{red!6}\textbf{3} & \textbf{MiniCPM-V 2.6} & \textbf{64.7} \\
            4 & VideoLLaMA2.1-AV & 58.8 \\
            5 & GPT-4o & 58.8 \\
            6 & Qwen2-VL+PC-DCoT & 56.2 \\
            7 & MiniCPM-V 2.6+PC-DCoT & 56.2 \\
            8 & InternVL2 & 52.9 \\
            9 & LLaVA-OneVision & 52.9 \\
            10 & GPT-4o+PC-DCoT & 50.0 \\
            11 & Aria & 47.1 \\
            12 & LLaVA-Video & 41.2 \\
     
        \end{tabular}
    }
    \subcaption{ \textit{Future Predictions}}
    \vspace{-2px}
\end{minipage}
\hfill
\begin{minipage}[t]{0.22\textwidth} 
    
    \centering
    \setlength{\tabcolsep}{3.0pt}
    \resizebox{1\linewidth}{!}{%
        \begin{tabular}{c|l|c}
            \textbf{Rank} & \multicolumn{1}{c|}{\textbf{Model}} & \textbf{Score} \\
            \Xhline{1.0pt}
                    \rowcolor{red!20}\faTrophy & \textbf{Qwen2-VL+PC-DCoT} & \textbf{69.1} \\
            \rowcolor{red!12}\textbf{2} & \textbf{InternVL2+PC-DCoT} & \textbf{68.1} \\
            \rowcolor{red!6}\textbf{3} & \textbf{LLaVA-OneVision} & \textbf{65.0} \\
            4 & GPT-4o+PC-DCoT & 62.8 \\
            5 & Qwen2-VL & 61.0 \\
            6 & MiniCPM-V 2.6+PC-DCoT & 60.6 \\
            7 & MiniCPM-V 2.6 & 59.0 \\
            8 & InternVL2 & 59.0 \\
            9 & GPT-4o & 59.0 \\
          
            10 & LLaVA-Video & 51.0 \\
            11 & Aria & 46.0 \\
            12 & VideoLLaMA2.1-AV & 45.0 \\
        \end{tabular}
    }
    \subcaption{ \textit{Current Interpretation}}
    \vspace{-2px}
\end{minipage}
\hfill
\begin{minipage}[t]{0.22\textwidth} 
    
    \centering
    \setlength{\tabcolsep}{3.0pt}
    \resizebox{1\linewidth}{!}{%
        \begin{tabular}{c|l|c}
            \textbf{Rank} & \multicolumn{1}{c|}{\textbf{Model}} & \textbf{Score} \\
            \Xhline{1.0pt}
                    \rowcolor{red!20}\faTrophy & \textbf{MiniCPM-V 2.6+PC-DCoT} & \textbf{73.0} \\
            \rowcolor{red!12}\textbf{2} & \textbf{InternVL2+PC-DCoT} & \textbf{68.0} \\
            \rowcolor{red!6}\textbf{3} & \textbf{GPT-4o+PC-DCoT} & \textbf{68.0} \\
            4 & Qwen2-VL+PC-DCoT & 64.0 \\
            5 & LLaVA-Video & 57.9 \\
            6 & GPT-4o & 56.1 \\
        
            7 & LLaVA-OneVision & 51.4 \\
            8 & InternVL2 & 49.5 \\
            9 & Qwen2-VL & 48.6 \\
            10 & MiniCPM-V 2.6 & 47.7 \\
            11 & Aria & 47.7 \\
            12 & VideoLLaMA2.1-AV & 43.9 \\
        \end{tabular}
    }
    \subcaption{ \textit{Figures Interactions}}
    \vspace{-2px}
\end{minipage}
\hfill
\begin{minipage}[t]{0.22\textwidth} 
    
    \centering
    \setlength{\tabcolsep}{3.0pt}
    \resizebox{1\linewidth}{!}{%
        \begin{tabular}{c|l|c}
            \textbf{Rank} & \multicolumn{1}{c|}{\textbf{Model}} & \textbf{Score} \\
            \Xhline{1.0pt}
                    \rowcolor{red!20}\faTrophy & \textbf{Qwen2-VL+PC-DCoT} & \textbf{81.3} \\
            \rowcolor{red!12}\textbf{2} & \textbf{GPT-4o+PC-DCoT} & \textbf{79.3} \\
            \rowcolor{red!6}\textbf{3} & \textbf{InternVL2+PC-DCoT} & \textbf{78.7} \\
            4 & MiniCPM-V 2.6+PC-DCoT & 76.0 \\
            5 & Qwen2-VL & 53.9 \\
            6 & GPT-4o & 53.2 \\
            7 & MiniCPM-V 2.6 & 48.7 \\
            8 & LLaVA-Video & 46.8 \\
       
            9 & VideoLLaMA2.1-AV & 44.2 \\
            10 & InternVL2 & 44.2 \\
            11 & Aria & 42.9 \\
            12 & LLaVA-OneVision & 42.9 \\
        \end{tabular}
    }
    \subcaption{ \textit{Figures Actions}}
    \vspace{-2px}
\end{minipage}

\vspace{0.1cm}

\begin{minipage}[t]{0.22\textwidth} 
    
    \centering
    \setlength{\tabcolsep}{3.0pt}
    \resizebox{1\linewidth}{!}{%
        \begin{tabular}{c|l|c}
            \textbf{Rank} & \multicolumn{1}{c|}{\textbf{Model}} & \textbf{Score} \\
            \Xhline{1.0pt}
                    \rowcolor{red!20}\faTrophy & \textbf{InternVL2+PC-DCoT} & \textbf{84.6} \\
            \rowcolor{red!12}\textbf{2} & \textbf{GPT-4o+PC-DCoT} & \textbf{82.1} \\
            \rowcolor{red!6}\textbf{3} & \textbf{MiniCPM-V 2.6+PC-DCoT} & \textbf{76.9} \\
            4 & Qwen2-VL+PC-DCoT & 74.4 \\
            5 & MiniCPM-V 2.6 & 58.5 \\
            6 & GPT-4o & 56.1 \\
            7 & Qwen2-VL & 51.2 \\
            8 & InternVL2 & 51.2 \\
            9 & VideoLLaMA2.1-AV & 39.0 \\
            10 & Aria & 39.0 \\
            11 & LLaVA-Video & 36.6 \\
           
            12 & LLaVA-OneVision & 31.7 \\
        \end{tabular}
    }
    \subcaption{ \textit{Characters Reference}}
    \vspace{-2px}
\end{minipage}
\hfill
\begin{minipage}[t]{0.22\textwidth} 
    
    \centering
    \setlength{\tabcolsep}{3.0pt}
    \resizebox{1\linewidth}{!}{%
        \begin{tabular}{c|l|c}
            \textbf{Rank} & \multicolumn{1}{c|}{\textbf{Model}} & \textbf{Score} \\
            \Xhline{1.0pt}
                    \rowcolor{red!20}\faTrophy & \textbf{InternVL2+PC-DCoT} & \textbf{88.0} \\
            \rowcolor{red!12}\textbf{2} & \textbf{GPT-4o+PC-DCoT} & \textbf{88.0} \\
            \rowcolor{red!6}\textbf{3} & \textbf{Qwen2-VL+PC-DCoT} & \textbf{84.0} \\
            4 & MiniCPM-V 2.6+PC-DCoT & 82.0 \\
            5 & Qwen2-VL & 67.9 \\
            6 & GPT-4o & 67.9 \\
            7 & InternVL2 & 64.2 \\
            8 & Aria & 62.3 \\
            9 & MiniCPM-V 2.6 & 54.7 \\
       
            10 & LLaVA-Video & 50.9 \\
            11 & LLaVA-OneVision & 50.9 \\
            12 & VideoLLaMA2.1-AV & 43.4 \\
        \end{tabular}
    }
    \subcaption{ \textit{Scene Transitions}}
    \vspace{-2px}
\end{minipage}
\hfill
\begin{minipage}[t]{0.22\textwidth} 
    
    \centering
    \setlength{\tabcolsep}{3.0pt}
    \resizebox{1\linewidth}{!}{%
        \begin{tabular}{c|l|c}
            \textbf{Rank} & \multicolumn{1}{c|}{\textbf{Model}} & \textbf{Score} \\
            \Xhline{1.0pt}
                    \rowcolor{red!20}\faTrophy & \textbf{Qwen2-VL} & \textbf{93.8} \\
            \rowcolor{red!12}\textbf{2} & \textbf{GPT-4o} & \textbf{93.8} \\
            \rowcolor{red!6}\textbf{3} & \textbf{MiniCPM-V 2.6+PC-DCoT} & \textbf{92.9} \\
            4 & Qwen2-VL+PC-DCoT & 85.7 \\
            5 & InternVL2+PC-DCoT & 78.6 \\
            6 & GPT-4o+PC-DCoT & 78.6 \\
            7 & MiniCPM-V 2.6 & 68.8 \\
            8 & Aria & 68.8 \\
            9 & InternVL2 & 62.5 \\
   
            10 & LLaVA-OneVision & 62.5 \\
            11 & VideoLLaMA2.1-AV & 56.2 \\
            12 & LLaVA-Video & 50.0 \\
        \end{tabular}
    }
    \subcaption{ \textit{Spatiotemporal Shifts}}
    \vspace{-2px}
\end{minipage}
\hfill
\begin{minipage}[t]{0.22\textwidth} 
    
    \centering
    \setlength{\tabcolsep}{3.0pt}
    \resizebox{1\linewidth}{!}{%
        \begin{tabular}{c|l|c}
            \textbf{Rank} & \multicolumn{1}{c|}{\textbf{Model}} & \textbf{Score} \\
            \Xhline{1.0pt}
                    \rowcolor{red!20}\faTrophy & \textbf{GPT-4o+PC-DCoT} & \textbf{75.3} \\
            \rowcolor{red!12}\textbf{2} & \textbf{MiniCPM-V 2.6+PC-DCoT} & \textbf{71.6} \\
            \rowcolor{red!6}\textbf{3} & \textbf{InternVL2+PC-DCoT} & \textbf{69.1} \\
            4 & Qwen2-VL+PC-DCoT & 64.2 \\
            5 & LLaVA-OneVision & 56.1 \\
            6 & LLaVA-Video & 54.9 \\
            7 & Qwen2-VL & 51.2 \\
            8 & Aria & 51.2 \\
            9 & InternVL2 & 51.2 \\
            10 & MiniCPM-V 2.6 & 48.8 \\
            11 & GPT-4o & 45.1 \\
            12 & VideoLLaMA2.1-AV & 43.9 \\
    
        \end{tabular}
    }
    \subcaption{ \textit{Object Interaction}}
    \vspace{-2px}
\end{minipage}

\vspace{0.1cm}

\vspace{-12px}
\caption{\textbf{Leaderboards of different Fine-Grained Task in {\BenchName}.}}
% \vspace{-0.3cm}
\label{tab:leaderboard}
\end{table*}

% prompt prompt_pc-dcot

% case data

\end{document}